\documentclass[10pt,twocolumn,letterpaper]{article}

\usepackage[pagenumbers]{wacv} % To force page numbers, e.g. for an arXiv version
\usepackage{booktabs}
\usepackage{multirow}
\usepackage{colortbl}
\usepackage{algorithm}
\usepackage{algpseudocode}
\usepackage{varwidth}
\makeatletter
\DeclareSymbolFont{extraup}{U}{zavm}{m}{n}
\DeclareMathSymbol{\newcheckmark}{\mathalpha}{extraup}{128}%uni2713
\DeclareMathSymbol{\newcrossmark}{\mathalpha}{extraup}{129}%uni2717
\makeatother
\newcommand\blfootnote[1]{%
  \begingroup
  \renewcommand\thefootnote{}\footnote{#1}%
  \addtocounter{footnote}{-1}%
  \endgroup
}
\definecolor{wacvblue}{rgb}{0.21,0.49,0.74}
\usepackage[pagebackref,breaklinks,colorlinks,allcolors=wacvblue]{hyperref}
\definecolor{rowcolor}{RGB}{205, 193, 255}
\definecolor{checkmarkgreen}{RGB}{0, 106, 103}
\definecolor{goodgreen}{RGB}{17, 139, 80}
\definecolor{comment}{RGB}{33, 28, 132}
\def\wacvPaperID{249} % *** Enter the WACV Paper ID here
\def\confName{WACV}
\def\confYear{2026}

\title{Grounding Degradations in Natural Language for All-In-One Video Restoration}

\author{Muhammad Kamran Janjua\(^{\clubsuit}\)\(^{*}\), Amirhosein Ghasemabadi\(^{\heartsuit}\)\(^{*}\), Kunlin Zhang\(^{\clubsuit}\), 
\\ Mohammad Salameh\(^{\clubsuit}\), Chao Gao\(^{\clubsuit}\), Di Niu\(^{\heartsuit}\)\\
\(^{\clubsuit}\)Huawei Technologies, Canada \\ \(^{\heartsuit}\)ECE Department, University of Alberta, Canada\\
{\tt\small muhammad.kamran.janjua@huawei.com, \{ghasemab,dniu\}@ualberta.ca}
}

\begin{document}
\maketitle
\begin{abstract}
In this work, we propose an all-in-one video restoration framework that grounds degradation-aware semantic context of video frames in natural language via foundation models, offering interpretable and flexible guidance. Unlike prior art, our method assumes no degradation knowledge in train or test time and learns an approximation to the grounded knowledge such that the foundation model can be safely disentangled during inference adding no extra cost. Further, we call for standardization of benchmarks in all-in-one video restoration, and propose two benchmarks in multi-degradation setting, three-task (3D) and four-task (4D), and two time-varying composite degradation benchmarks; one of the latter being our proposed dataset with varying snow intensity, simulating how weather degradations affect videos naturally. We compare our method with prior works and report state-of-the-art performance on all benchmarks.\blfootnote{\(^{*}\) indicates equal contribution}
\end{abstract}
    
\section{Introduction}
\label{sec:intro}

Video restoration aims to restore a given  degraded, low-quality video~\citep{rota2023video,liu2022video}. Traditional work in this area tends to address each type of degradation separately~\citep{liang2022recurrent,liang2024vrt,tassano2020fastdvdnet,li2023simple,ghasemabadilearning}, where a stand-alone parameterized model learns to reverse that degradation. The generalization of restoration procedure to a mixture of degradations, i.e., learning a single model for multiple degradations, is referred to by an umbrella term \textit{all-in-one restoration} in the literature~\citep{li2022all,potlapalli2023promptir,valanarasu2022transweather}. The goal thereof being to recover a high-quality video by reversing several degradations at once.

Majority of the mainstream methods attempting to solve the said problem fall into one of three categories: implicit (aka blackbox) prompt, explicit (aka whitebox) prompt, or discriminative. Implicit prompt methods learn degradation priors from the input data to condition and guide the reconstruction phase~\citep{potlapalli2023promptir,li2023prompt,zhaoavernet,valanarasu2022transweather}. These methods, however, lack interpretability and offer limited conditioning and control, making it increasingly difficult to understand what the prompt has learned. Explicit prompt methods leverage an external model-based knowledge base, typically multimodal large language models (MLLMs), to condition the reconstruction phase~\citep{conde2024high,luo2023controlling,yan2023textual,yang2023language,jin2024llmra}. Their tight coupling with external models reduces computational efficiency, and repeatedly querying an MLLM per video frame in inference is expensive and often impractical. Discriminative methods rely on contrastive learning in the latent space to learn degradation-specific representations~\citep{li2022all,yang2023video}. However, just like explicit prompt methods, they assume access to degradation information, but additionally require that only one degradation affects any given frame, which is an unnatural assumption in degraded videos. Due to this limiting assumption, they cannot reliably function in composite degradation settings wherein multiple degradations corrupt a single frame. A brief summary of these methods is presented in~\cref{tab:method_comparison}.

Further, in all-in-one image restoration, there has been consistent work~\citep{jin2024llmra,yan2023textual,li2023prompt,yang2023language,conde2024high,li2022all,potlapalli2023promptir,valanarasu2022transweather,cui2024adair,luo2023controlling,ma2023prores,lin2023improving}, and the literature has well-defined benchmarks, both in terms of the tasks and datasets for the problem. However, there exists a gap in addressing the all-in-one restoration problem in the video restoration literature, and the progress is siloed. Nonetheless, a few disparate attempts have been made with each method tackling a different combination of degradations for the problem, all the while reporting performance on distinct video datasets~\citep{hui2024vjt,shekarforoush2023dual,cheng2023cross,yang2023video}.

In this work, we offer a fresh perspective on how multimodal language models can aid challenges in video understanding, with a particular focus on video restoration. To this end, we propose a no bells-and-whistles framework, which we call \textsc{Ronin}, that grounds the degradation-aware semantic context of video frames in natural language without the need for explicit a priori degradation information or deploying MLLMs in inference time. To unify the heterogeneity in all-in-one video restoration methods and standardize benchmarks for future research to build on, we propose two benchmarks in multi-degradation setting, and two time-varying composite degradation benchmarks. In the latter case, we introduce a new benchmark that extends time-varying unknown degradations to weather, particularly snow. Our contributions are listed as follows:
\begin{itemize}
    \item We introduce a novel method to g\textbf{RO}u\textbf{N}d the degradat\textbf{I}ons in la\textbf{N}guage, termed as \textsc{Ronin}, and condition the all-in-one restoration procedure on the degradation-aware semantic context of video frames.
    \item \textsc{Ronin} grounds each degraded frame in natural language and learns to distill this information throughout the training to function standalone in inference.
    \item We extend the time-varying unknown degradation (TUD) setting to weather and introduce the \texttt{SnowyScenes} benchmark with varying snow intensity across videos.
    \item We standardize the all-in-one video restoration literature and propose two benchmarks in multi-degradation setting, namely \(3\)D and \(4\)D, along with two composite degradation benchmarks, time-varying unknown degradations, TUD~\citep{zhaoavernet}, and \texttt{SnowyScenes}.
\end{itemize}
\section{Related Work}
\label{sec:relatedwork}

%% Updated Starts
Image and video restoration problems are well-studied in the literature~\citep{chen2022simple,zamir2022restormer,liang2021swinir,ghasemabadi2024cascadedgaze,liang2022recurrent,liang2024vrt,huang2022neural,li2023simple,ghasemabadilearning}. Recently, there has been a surge in methods learning a single parameterized model to restore several degradations simultaneously. This approach is referred to as all-in-one restoration. Various methods for all-in-one restoration have been proposed, primarily using backbone architectures constructed in either columnar~\citep{liang2021swinir} or UNet~\citep{ronneberger2015u} fashion.

\paragraph{All-In-One Image Restoration} AirNet~\citep{li2022all} established an early benchmark by using a contrastive degradation encoder, while TransWeather~\citep{valanarasu2022transweather} proposed to incorporate weather-specific queries within a Transformer framework. Building on these ideas, works such as PromptIR~\citep{potlapalli2023promptir} and Prompt-In-Prompt~\citep{li2023prompt} proposed blackbox prompt methods. On the other hand, language-guided whitebox prompt approaches such as InstructIR~\citep{conde2024high}, LLMRA~\citep{jin2024llmra}, LanguageWeather~\citep{yang2023language}, and TextIR~\citep{yan2023textual} inject human-aligned instructions or textual features into the restoration method. In blackbox prompt methods, the prompts are not interpretable, making it increasingly difficult to understand what the prompt has learned\footnote{Although some basic understanding of their discriminative behavior is possible through visualization.}. While in the case of whitebox prompt methods, the language model or vision-language model can not be disentangled from the underlying restoration method in inference, which increases overall computational costs and hinders deployability.

\paragraph{All-In-One Video Restoration} All image restoration methods discussed above are comparable to each other since they are evaluated consistently on similar all-in-one restoration datasets and tasks. However, the progress in all-in-one video restoration is siloed, and the attempts made in the literature are disparate in nature. Methods like VJT~\citep{hui2024vjt} and CDUN~\citep{cheng2023cross} extend the all-in-one restoration framework to handle diverse degradations in videos but rely on proprietary datasets that are not publicly available, 
making it challenging for subsequent methods to benchmark their performance. 
More recent contributions, such as ViWS-Net~\citep{yang2023video} and AverNet~\citep{zhaoavernet} address weather-specific and time-varying degradations, highlighting ongoing challenges in creating a standardized all-in-one video restoration paradigm. A comprehensive literature review is deferred to the appendix.

\begin{table}[]
\centering
\scalebox{0.78}{
\begin{tabular}{@{}lccccc@{}}
\toprule
\multirow{2}{*}{\textbf{Method}} & \multicolumn{2}{c}{\begin{tabular}[c]{@{}c@{}}\textbf{No Prior}\\ \textbf{Degradation}\\ \textbf{Knowledge}\end{tabular}} & \multirow{2}{*}{\begin{tabular}[c]{@{}c@{}}\textbf{Natural}\\ \textbf{Language} \\ \textbf{Prompt}\end{tabular}} & \multirow{2}{*}{\begin{tabular}[c]{@{}c@{}}\textbf{No} \\ \textbf{Additional}\\ \textbf{Network}\end{tabular}} & \multirow{2}{*}{\begin{tabular}[c]{@{}c@{}}\textbf{Params}\\ (M) \(\downarrow\) \end{tabular}}  \\ \cmidrule(lr){2-3}
 & \multicolumn{1}{l}{\begin{tabular}[c]{@{}l@{}}Train\\\end{tabular}} & \multicolumn{1}{l}{\begin{tabular}[c]{@{}l@{}}Inference\\\end{tabular}} &  &  &  \\ \midrule
AirNet~\citep{li2022all} & \(\mathcolor{red}{\newcrossmark}\) & \Large \textcolor{checkmarkgreen}{\textbf{\checkmark}} & \(\mathcolor{red}{\newcrossmark}\) & \Large \textcolor{checkmarkgreen}{\textbf{\checkmark}} & \(7.6\)M \\
PromptIR~\citep{potlapalli2023promptir} & \Large \textcolor{checkmarkgreen}{\textbf{\checkmark}} & \Large \textcolor{checkmarkgreen}{\textbf{\checkmark}} & \(\mathcolor{red}{\newcrossmark}\) & \Large \textcolor{checkmarkgreen}{\textbf{\checkmark}} & \(35.59\) M \\
InstructIR~\citep{conde2024high} & \(\mathcolor{red}{\newcrossmark}\) & \(\mathcolor{red}{\newcrossmark}\) & \Large \textcolor{checkmarkgreen}{\textbf{\checkmark}} & \(\mathcolor{red}{\newcrossmark}\) & \(73.95\) M \\
ViWSNet~\citep{yang2023video} & \(\mathcolor{red}{\newcrossmark}\) & \Large \textcolor{checkmarkgreen}{\textbf{\checkmark}} & \(\mathcolor{red}{\newcrossmark}\) & \(\mathcolor{red}{\newcrossmark}\) & \(57.82\) M \\
AverNet~\citep{zhaoavernet} & \Large \textcolor{checkmarkgreen}{\textbf{\checkmark}} & \Large \textcolor{checkmarkgreen}{\textbf{\checkmark}} & \(\mathcolor{red}{\newcrossmark}\) & \(\mathcolor{red}{\newcrossmark}\) & \(41.35\) M\(^{\mathcolor{red}{*}}\) \\ \midrule
\rowcolor{rowcolor} \textsc{\textbf{Ronin}} & \Large \textcolor{checkmarkgreen}{\textbf{\checkmark}} & \Large \textcolor{checkmarkgreen}{\textbf{\checkmark}} & \Large \textcolor{checkmarkgreen}{\textbf{\checkmark}} & \Large \textcolor{checkmarkgreen}{\textbf{\checkmark}} & \(57.0\) M \\ \bottomrule
\end{tabular}
}
\caption{\textbf{Summary of Prior Methods.} We summarize the prior methods in terms of their conditioning style (if the prompt is interpretable aka whitebox or not), the need for additional modules (such as optical flow for motion compensation in AverNet~\citep{zhaoavernet} or text-encoder in InstructIR~\citep{conde2024high}), and the assumption of degradation type as a prior during training or inference. We also present number of parameters of each method. Note that \(^{\mathcolor{red}{*}}\) indicates that the parameters of optical flow model were not included. Our method, \textsc{Ronin}, is prior-free, injects interpretable whitebox prompts, and requires no additional network to restore videos.}
\label{tab:method_comparison}
\end{table}

\begin{figure*}[!t]
    \centering
    \includegraphics[width=\linewidth]{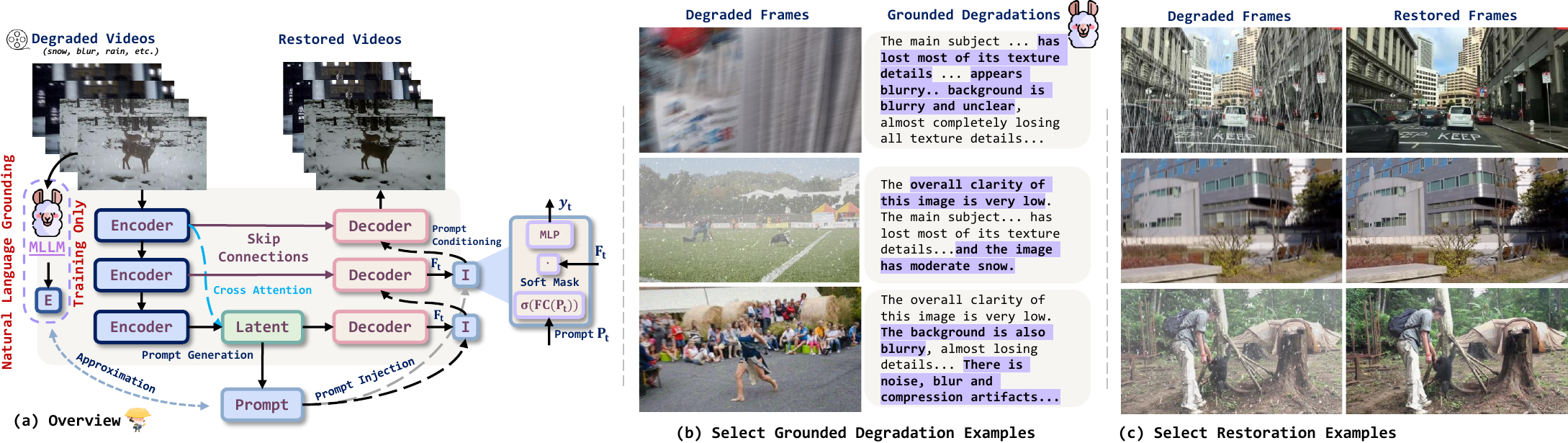}
    \caption{\textbf{Bird's-Eye View of \textsc{Ronin} and Assorted Examples.} We visualize \textsc{Ronin}'s architecture in (a) along with a few grounded degradation examples in (b) and restoration results in (c). The grounded degradations in (b) are \colorbox{rowcolor}{highlighted} to emphasize the text that describes the degradations and quality of the image/frame. The restoration frames in (c) are taken from video deraining, video deblurring, and video desnowing tasks, respectively.}
    \label{fig:mainfig}
\end{figure*}

\section{Methodology}
\label{sec:method}
We consider a low-quality video \(\mathbf{V}^{\text{LQ}} \in \mathbb{R}^{T \times H \times W \times C}\) afflicted by unknown degradations \(\{d_{0}, d_{1}, ..., d_{n}\} \in \mathcal{D}\), where \(T, H, W, C\) denote temporal, height, width and channel dimensions, respectively. In all-in-one video restoration, the goal is to learn a single model \(M_\theta\), parameterized by \(\theta\), to reverse various degradations and obtain a high-quality video \(\mathbf{V}^{\text{HQ}} \in \mathbb{R}^{T \times H \times W \times C}\). 
%Different from traditional video restoration, in all-in-one, degradations either vary with time in a single video~\citep{zhaoavernet}, are composite, i.e., various degradations afflict the same video, or a single network is responsible for multiple degradations in different videos. 
Unlike traditional video restoration methods, the degradations in all-in-one restoration may vary over time within a single video~\citep{zhaoavernet}, may be composite (with multiple degradations affecting the same video), or a single network may handle multiple types of degradations across different videos. 
We approach the problem of all-in-one video restoration through conditioning the restoration backbone with explicit whitebox prompt injections. Specifically, we ground the degradations affecting each frame in natural language and inject this as prior knowledge in the restoration network, assuming no knowledge of degradations at train or test time.

%Since our method works online, we treat each video as streaming video.
We treat each video as streaming video since our method operates online.
%Our underlying restoration network is built on top of Turtle~\citep{ghasemabadilearning} which is a U-Net~\citep{ronneberger2015u} style architecture that processes streaming videos. 
Our restoration network is based on the Turtle~\citep{ghasemabadilearning} architecture which is a U-Net~\citep{ronneberger2015u} style architecture that processes streaming videos. 
Given a frame at timestep \(t\), Turtle models the causal relationship \(p(\mathbf{y}_{t}|\mathbf{F}_{t}, \mathbf{H}_{t})\), where \(\mathbf{y}_{t}\) is the output, \(\mathbf{F}_{t}\) is the feature map of the input frame at time \(t\), and \(\mathbf{H}_{t}\) is the history of corresponding features maps from the previous frames. In this work, we modify this formulation for two decoder blocks as illustrated in~\cref{fig:mainfig}(a), and inject prompts generated from the latent features. Specifically,  \textsc{Ronin} learns to model \(p(\mathbf{y}_{t}|\mathbf{F}_{t}, \mathbf{H}_{t}, \mathbf{P}_{t})\) where \(\mathbf{P}_{t}\) is the prompt.

% Following the original work~\citep{ghasemabadilearning}, we train \textsc{Ronin} with standard loss \(L_{1}\) loss function: \(\mathcal{L} = \frac{1}{N}\sum_{i=1}^{N} \|\mathbf{V}^{\text{GT}} - \mathbf{V}^{\text{HQ}}\|\) for all experiments reported in this work.

\subsection{Grounding Degradations in Language}
We assume access to a multimodal large language model (MLLM) capable of taking a frame as input and describing the degradations affecting the frame and its content. We find that Q-Instruct~\citep{wu2023q}, which is built on LLaVA~\citep{liu2023llava}, to be sufficient for this purpose. Given a frame at timestep \(t\), we query the MLLM and prompt it to describe the image quality by feeding it the degraded frame along with the prompt: \textit{‘Rate the quality of the image. Think step by step.’} 

% \footnote{Our goal is to generate a description of the degradation, not a detailed assessment of quality or the level of degradation; therefore, we do not further fine-tune Q-Instruct.} 
%Given a frame at timestep \(t\), we query the MLLM and prompt it to describe the quality of the image, specifically, we feed the degraded frame and prompt \textit{`Rate the quality of the image. Think step by step'}. 

\begin{figure}
    \centering
    \includegraphics[width=\linewidth]{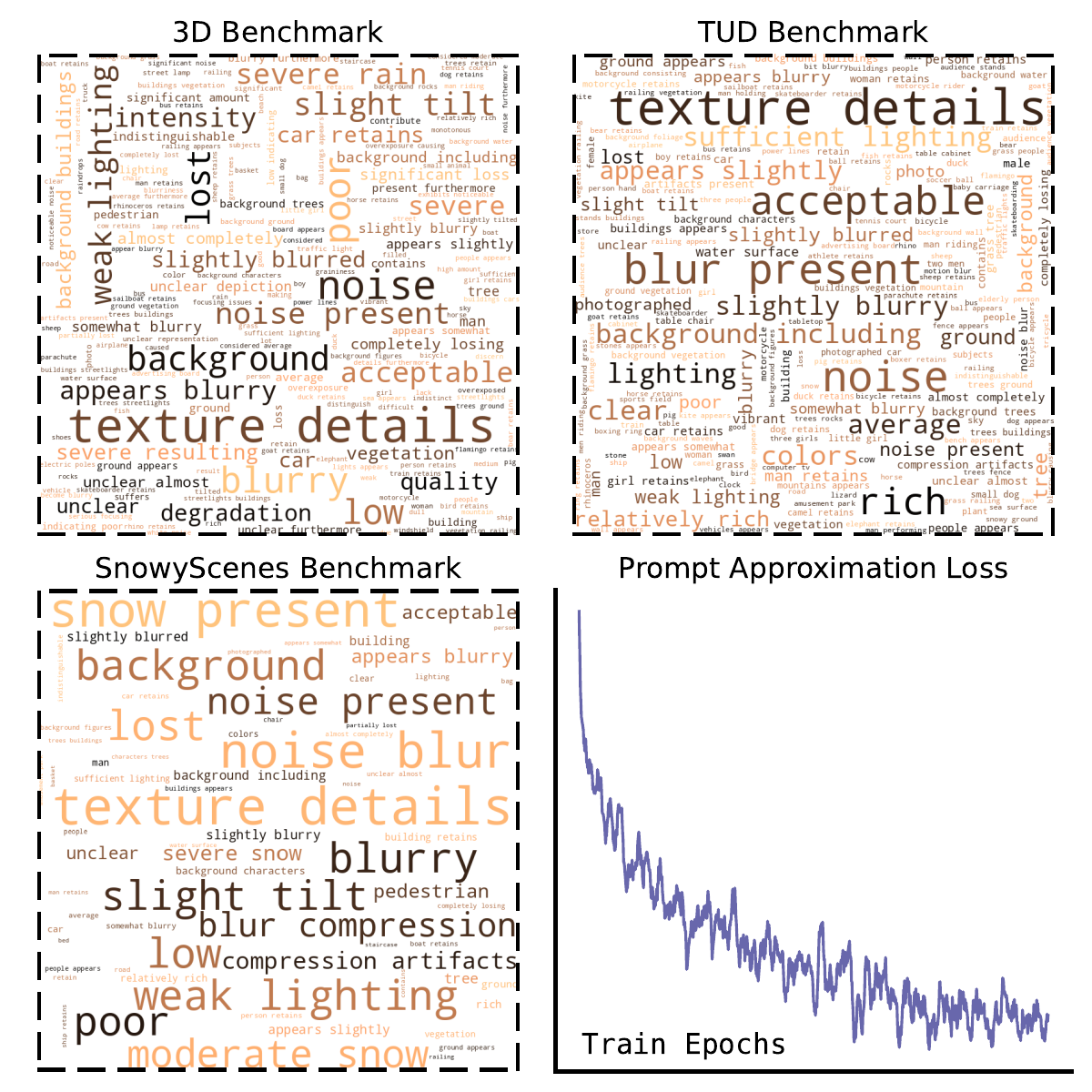}
    \caption{\textbf{Word Cloud of Different Benchmarks.} We visualize word cloud of per-frame language descriptions generated from Q-Instruct~\citep{wu2023q} for three benchmarks, i.e., \(3\)D, TUD~\citep{zhaoavernet} and our proposed \texttt{SnowyScenes}. We also plot the prompt approximation loss during training (bottom right) to verify that the optimization procedure converges.}
    \label{fig:wordcloud}
\end{figure}

The output content and degradation description is then fed to a language encoder to get the vector embeddings. Some examples of the descriptions are presented in~\cref{fig:mainfig}(b) and in appendix. In practice, we generate language descriptions per frame and its corresponding vector embeddings from the language encoder offline,
storing them for later querying during training. We visualize the language descriptions in a word cloud in~\cref{fig:wordcloud} for three benchmark datasets we consider in this work. 
For instance, in the \(3\)D benchmark, there are noise, rain, and blur degradations, and the word cloud effectively represents all three types (top right).
%Consider how in \(3\)D benchmark, there are noise, rain and blur degradations, and its word cloud is representative of all three degradations (top right). 
% the snow intensity varies which is indicated by words \textit{moderate snow} or \textit{severe snow} in the word cloud (bottom left) indicating that the prompts appropriately ground per-frame degradations in natural language.
Similarly, in our proposed \texttt{SnowyScenes} benchmark, the variation in snow intensity is reflected in the word cloud with terms such as \textit{moderate snow} or \textit{severe snow} (bottom left), indicating that the prompts adequately ground per-frame degradations in natural language. We use BGE-Micro-v2 to generate vector embeddings since it is a lightweight text encoder.
\footnote{\href{https://huggingface.co/TaylorAI/bge-micro-v2}{https://huggingface.co/TaylorAI/bge-micro-v2}} However, unlike~\citep{conde2024high}, we do not fine-tune the language encoder or employ any classification loss on the text embeddings. This is because it requires access to degradation type and significantly suffers in time-varying or composite type degradations wherein multiple degradations afflict the same video, and is often a natural setting given how degradations occur in videos. 
% Our approach is advantageous because it does not require information about the type of degradation. Also, it performs better in scenarios with time-varying or composite degradations, where multiple degradations affect the same video—situations that are quite common in reality. 
%Note that such a setup affords 
Our setup offers three major benefits (i) no assumption or prior on the degradation is required since the MLLM automatically assesses and generates appropriate degradation descriptions, (ii) interpretable prompts allow nuances in conditioning and guidance since plain instructions are rigid and cannot adequately describe composite degradations, and (iii) per-frame prompts allows processing streaming videos wherein a single unique description is tailored to each frame. We discuss designing prompt template and present more examples in the appendix.
% for a unique description tailored to each frame. 

\subsection{Prompt Generation and Injection}
In \textsc{Ronin}, prompt component consists of a set of learnable parameters that are generated from the incoming frame features at the latent stage.
These parameters create an embedding of the language description related to the degradation present in the input frame. 
%to form the information that embeds the language description of the degradation afflicting the input frame. 
Since the restoration network is a U-Net, the feature map is compressed in the encoder stages, and is inflated back in the decoder stage. At the latent stage, most degradation information has been removed, and only essential input information necessary for reconstruction remains. To allow some degradation information, we introduce some information from the first encoder stage through cross-attention back into the latent feature map that generates the prompt, see~\cref{fig:mainfig}. This encourages efficient information mixing to generate a compact prompt that already embeds the degradation information inherent in the frame, and adapts itself to approximate its language grounded representation. It is intuitive to learn the prompts dynamically and let them be dependent on the input since each condition differs from the others in videos (e.g., degradation may change, content may change, etc.).

\paragraph{Prompt Generation} 
% In the same spirit as~\citep{potlapalli2023promptir,li2023prompt,zhaoavernet}, 
To generate a prompt, we first compute the average of the input feature map across the spatial dimension, i.e., we perform Global Average Pooling (GAP), to obtain a feature vector \(\mathbf{v}_{t} \in \mathbb{R}^{b \times C}\), where \(b\) is the batch size and \(C\) denotes the number of channels. 
We then project this vector \(\mathbf{v}_{t}\) along the same dimensions as the text encoder followed by a GELU~\citep{hendrycks2016gaussian} non-linearity. To allow the prompt to adjust as it learns to approximate the language description during training, we project it through another linear layer but maintain the dimensions. %, i.e., 
This process can be expressed as follows:
\begin{equation}
\label{eq:promptgen}
    \mathbf{P}_{t} = \texttt{FC}(\texttt{GELU}(\texttt{FC}(\texttt{GAP}(\mathbf{F}_{t})) \in \mathbb{R}^{b \times d},
\end{equation}
where \(d\) is the output embedding dimension from the text encoder, and \(\mathbf{F}_{t}\) is the feature map taken from the latent stage. This is beneficial since the spatial size of the feature map at latent stage is minimum with the highest number of channels, allowing for comprehensive information flow.
% Our method achieves comprehensive information flow since the spatial size of the feature map at latent stage is minimum with highest number of channels.

\paragraph{Prompt Injection} Given the generated prompt \(\mathbf{P}_{t}\), we then inject it in the last two decoder stages such that the restored output is modulated based on the language guidance, see~\cref{fig:mainfig}. Let \(\mathbf{F}_{t}^{[l-1]}\) denote the output from the previous layer, then the prompt injection procedure learns a soft-mask from the prompt \(\mathbf{P}_{t}\) to choose the features that are relevant to the task in consideration. Specifically, we first project the prompt \(\mathbf{P}_{t}\) through a linear layer followed by the sigmoid non-linearity to generate a per-channel soft-mask i.e., \(\sigma(\texttt{FC}(\mathbf{P}_{t}))\), where \(\sigma\) is the sigmoid function. This mask is then applied to the feature map from the previous layer as
\begin{equation}
\label{eq:promptinj}
    \mathbf{F}_{t}^{\mathbf{P}_{t}} = \mathbf{F}_{t}^{[l-1]} \odot \sigma(\texttt{FC}(\mathbf{P}_{t})),
\end{equation}
where \(\odot\) denotes multiplication operation and \(\mathbf{F}_{t}^{\mathbf{P}_{t}}\) denotes the prompt conditioned representation. The output is then fed to a simple MLP and is passed to the next decoder stage. This procedure is similar in spirit to several prompt injection modules whose goal is to combine the input representations with a prompt~\citep{conde2024high,potlapalli2023promptir,li2023prompt} or even just to modulate channels~\citep{he2019modulating,strezoski2019many,ghasemabadilearning}.

\begin{table*}[!t]
\centering
\scalebox{0.85}{
\begin{tabular}{@{}clcccccccccc@{}}
\toprule
\multirow{2}{*}{\textbf{Setting}} & \multirow{2}{*}{\textbf{Method}} & \multicolumn{2}{c}{\begin{tabular}[c]{@{}c@{}}\textbf{Deblur}\\ (GoPro~\citep{Nah_2017_CVPR})\end{tabular}} & \multicolumn{2}{c}{\begin{tabular}[c]{@{}c@{}}\textbf{Denoise}\\ (DAVIS~\citep{Pont-Tuset_arXiv_2017})\end{tabular}} & \multicolumn{2}{c}{\begin{tabular}[c]{@{}c@{}}\textbf{Derain}\\ (VRDS~\citep{wu2023mask})\end{tabular}} & \multicolumn{2}{c}{\begin{tabular}[c]{@{}c@{}}\textbf{Desnow}\\ (RVSD~\citep{chensnow})\end{tabular}} & \multicolumn{2}{c}{\textbf{Average}} \\ \cmidrule(l){3-12} 
 &  & \textbf{PSNR} \(\uparrow\) & \textbf{SSIM} \(\uparrow\) & \textbf{PSNR} \(\uparrow\) & \textbf{SSIM} \(\uparrow\) & \textbf{PSNR} \(\uparrow\) & \textbf{SSIM} \(\uparrow\) & \textbf{PSNR} \(\uparrow\) & \textbf{SSIM} \(\uparrow\) & \textbf{PSNR} \(\uparrow\) & \textbf{SSIM} \(\uparrow\) \\ \midrule
\multirow{6}{*}{\(\mathbf{3}\)\textbf{D}} & Restormer~\citep{zamir2022restormer} & \(31.1653\) & \(0.9462\) & \(31.3816\) & \(0.9193\) & \(31.1068\) & \(0.9555\) & \multicolumn{2}{c}{\multirow{2}{*}{}} & \(31.2179\) & \(0.9403\) \\
 & InstructIR~\citep{conde2024high} & \(30.9331\) & \(0.9439\) & \(31.2521\) & \(0.9158\) & \(31.0966\) & \(0.9547\) & \multicolumn{2}{c}{} & \(31.0939\) & \(0.9381\) \\
 & PromptIR~\citep{potlapalli2023promptir} & \(31.2833\) & \(0.9474\) & \(31.3529\) & \(0.9182\) & \(31.1776\) & \(0.9559\) & \multicolumn{2}{c}{\textbf{N/A}} & \(31.2713\) & \(0.9405\) \\
 & ViWSNet~\citep{yang2023video} & \(27.8949\) & \(0.8949\) & \(29.9601\) & \(0.8863\) & \(28.5579\) & \(0.9234\) & \multicolumn{2}{c}{\multirow{3}{*}{}} & \(27.6298\) & \(0.8250\) \\
 & AverNet~\citep{zhaoavernet} & \(30.8064\) & \(0.9157\) & \(25.2306\) & \(0.4934\) & \(\mathbf{32.8695}\) & \(0.9441\) & \multicolumn{2}{c}{} & \(29.6355\) & \(0.7844\) \\ \midrule
 \rowcolor{rowcolor} & \textsc{\textbf{Ronin}} & \(\mathbf{32.7327}\) & \(\mathbf{0.9605}\) & \(\mathbf{31.6539}\) & \(\mathbf{0.9220}\) & \(32.7224\) & \(\mathbf{0.9656}\) & \multicolumn{2}{c}{} & \(\mathbf{32.3696}\) & \(\mathbf{0.9493}\) \\ \midrule
\multirow{5}{*}{\(\mathbf{4}\)\textbf{D}} & Restormer~\citep{zamir2022restormer} & \(29.6629\) & \(0.9286\) & \(31.0225\) & \(0.9117\) & \(29.8737\) & \(0.9437\) & \(25.9196\) & \(0.9263\) & \(29.1196\) & \(0.9275\) \\
 & InstructIR~\citep{conde2024high} & \(29.4654\) & \(0.9260\) & \(31.0074\) & \(0.9125\) & \(29.8215\) & \(0.9442\) & \(24.8697\) & \(0.9163\) & \(28.7910\) & \(0.9247\) \\
 & PromptIR~\citep{potlapalli2023promptir} & \(29.7082\) & \(0.9296\) & \(31.0868\) & \(0.9130\) & \(30.2119\) & \(0.9481\) & \(\mathbf{26.1032}\) & \(\mathbf{0.9278}\) & \(29.2775\) & \(0.9296\) \\
 & ViWSNet~\citep{yang2023video} & \(27.2592\) & \(0.8821\) & \(29.6782\) & \(0.8853\) & \(28.1486\) & \(0.9185\) & \(24.8427\) & \(0.9028\) & \(27.4806\) & \(0.8972\) \\ \midrule
\rowcolor{rowcolor} & \textsc{\textbf{Ronin}} & \(\mathbf{30.7186}\) & \(\mathbf{0.9417}\) & \(\mathbf{31.2230}\) & \(\mathbf{0.9160}\) & \(\mathbf{31.1688}\) & \(\mathbf{0.9544}\) & \(25.9538\) & \(0.9237\) & \(\mathbf{29.7660}\) & \(\mathbf{0.9339}\) \\ \bottomrule
\end{tabular}
}
\caption{\textbf{3D and 4D Benchmark Results.} Quantitative results (PSNR and SSIM) on the \(3\)D and \(4\)D benchmarks comparing all-in-one restoration methods with \textsc{Ronin}.}
\label{tab:3d_4d_results}
\end{table*}

\subsection{Prompt Approximation} To make sure the prompt is meaningful without directly incorporating the MLLM/VLM~\citep{luo2023controlling} or a text encoder~\citep{conde2024high} in inference, we propose to approximate the relevant embeddings from the text encoder during training. We impose an optimization objective which in addition to restoring the frame, also penalizes if the generated prompt \(\mathbf{P}_{t}\) is not aligned with the text representation from the encoder. Let \(\mathbf{e}_{t}(C_{t})\) denote the text encoder representation for some grounded context (language description) \(C_{t}\) taken from the MLLM for a given frame. Then, a straight-forward \(L_{1}\) objective suffices to enforce that \(\mathbf{P}_{t} \approx \mathbf{e}_{t}(C_{t})\), and since the prompt is generated from the latent representation which is unique per sample, it does not collapse to an average text encoder representation. We find that empirically this works well, see~\cref{fig:wordcloud} where we visualize the prompt loss during training (bottom right). The overall optimization objective of \textsc{Ronin} is then given as follows
\begin{equation}
\label{eq:opt_obj}
    \mathcal{L} = \underbrace{\lambda_{1} \frac{1}{N}\sum^{N} \|\mathbf{V}^{\text{GT}} - \mathbf{V}^{\text{HQ}}\|}_{\text{Restoration Loss}} + \underbrace{\lambda_{2} \frac{1}{N}\sum^{N} \|e_{t}(C_{t}) - \mathbf{P}_{t}\|}_{\text{Prompt Approximation Loss}},
\end{equation}
where \(\lambda_{1}\) and \(\lambda_{2}\) are balancing factors and we set \(\lambda_{1} = 1.0\) and \(\lambda_{2} = 0.01\). Intuitively, the prompt approximation objective can be thought of as distilling the necessary degradation related information from the text encoder, or the grounded context from the MLLM, into the prompt generation module of \textsc{Ronin}.
\section{Experiments}
\label{sec:experiments}

We follow the standard experimental settings outlined in~\citep{ghasemabadilearning} to train \textsc{Ronin} for all the experiments reported in this manuscript. We train \textsc{Ronin} with Adam optimizer~\citep{kingma2014adam} and default beta values. The initial learning rate is set to \(4e^{-4}\) and is decayed to \(1e^{-7}\) throughout the training procedure through the cosine annealing strategy~\citep{loshchilov2016sgdr}. All of our models are implemented in the PyTorch library, and are trained on \(8\) NVIDIA Tesla v100 \(32\) GB GPUs for \(200\)k iterations; in the case of the TUD benchmark, we train our model for \(300\)k iterations. We query Q-Instruct~\citep{wu2023q} for each frame offline and extract the embeddings from the text encoder to store it in a dictionary which is queried during training. Note that both Q-Instruct~\cite{wu2023q} and the text encoder are not required in inference. Further, we assume no a priori degradation for all the tasks. We apply basic data augmentation techniques, including horizontal-vertical flips and 90-degree rotations. Following the video restoration literature, we use Peak Signal-to-Noise Ratio (PSNR) and Structural Similarity Index (SSIM)~\citep{wang2004image} distortion metrics to report quantitative performance. For qualitative evaluation, we present visual outputs for each task and compare them with the results obtained from previous methods.

% , but adapt it to take videos as input instead of images\footnote{We feed two frames (at timestep \(t-1\) and \(t\)) as input.}
\paragraph{Compared Methods} Given the lack of methods in all-in-one video restoration, we compare \textsc{Ronin} to one image restoration method Restormer~\cite{zamir2022restormer} and three representative all-in-one image restoration methods i.e., AirNet~\citep{li2022all}, PromptIR~\citep{potlapalli2023promptir}, and InstructIR~\citep{conde2024high}. We also consider two video restoration methods ViWSNet~\citep{yang2023video}, which is a method for restoring videos from adverse weather conditions, and AverNet~\citep{zhaoavernet}, which is a method for restoring time-varying unknown degradations in videos. Our baseline selection was based on each method’s approach, i.e., implicit blackbox prompts in PromptIR/AverNet, whitebox prompts in InstructIR, and contrastive learning in AirNet/ViWSNet, community usage, and open-source availability. For all the experiments, we follow the original code-bases of each of the said methods released by their respective authors and train and evaluate on our benchmarks.

\begin{figure*}[!t]
    \centering
    \includegraphics[width=\linewidth]{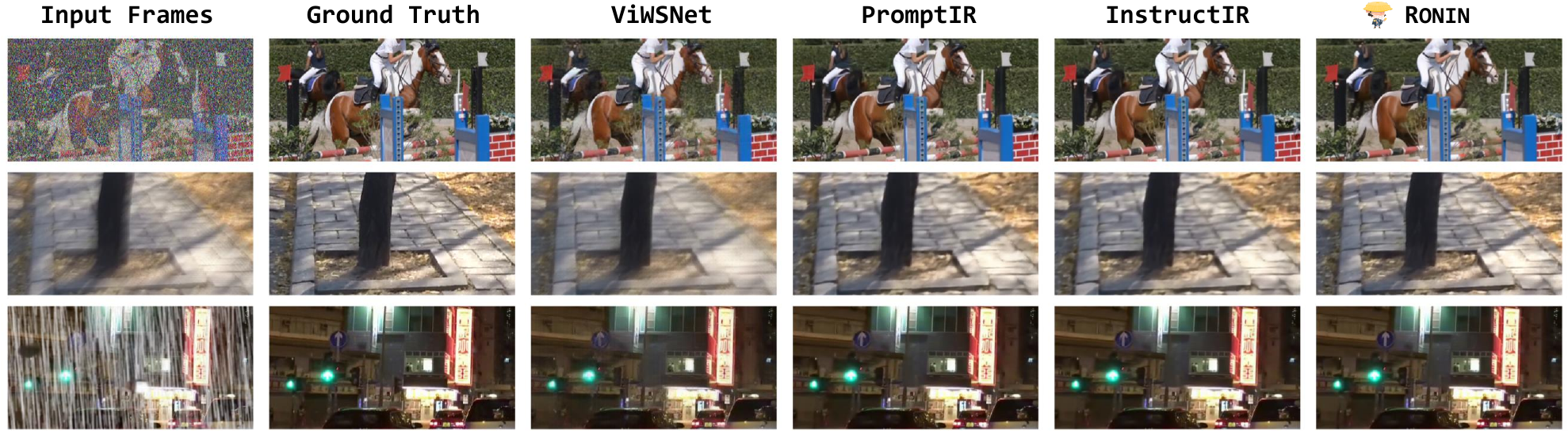}
    \caption{\textbf{Visual Results on \(\mathbf{3}\)D Benchmark.} We qualitatively compare three prior methods with \textsc{Ronin} on all tasks of the \(3\)D benchmark. The first row contains frame crops from denoising video, while the second and third row contain frames crops from deblurring and deraining videos, respectively. Notice how \textsc{Ronin}'s outputs are visually pleasing e.g., the person in the back on the horse and the folded leg of the brown horse in the denoising video, the stone texture in the deblurring video and the green arrow sign board in the deraining video. All of these regions in other methods' outputs are smeared. Best viewed zoomed-in.}
    \label{fig:3dvisuals}
\end{figure*}

\begin{table*}[!t]
\centering
\scalebox{0.8}{
\begin{tabular}{@{}lcccccccccccc@{}}
\toprule
\multirow{3}{*}{\textbf{Method}} & \multicolumn{6}{c}{\textbf{DAVIS}~\citep{Pont-Tuset_arXiv_2017}} & \multicolumn{6}{c}{\textbf{Set8}~\citep{tassano2019dvdnet}} \\ \cmidrule(l){2-13} 
 & \multicolumn{2}{c}{\(\mathbf{t = 6}\)} & \multicolumn{2}{c}{\(\mathbf{t = 12}\)} & \multicolumn{2}{c}{\(\mathbf{t = 24}\)} & \multicolumn{2}{c}{\(\mathbf{t = 6}\)} & \multicolumn{2}{c}{\(\mathbf{t = 12}\)} & \multicolumn{2}{c}{\(\mathbf{t = 24}\)} \\ \cmidrule(l){2-13} 
 & \textbf{PSNR} \(\uparrow\) & \textbf{SSIM} \(\uparrow\) & \textbf{PSNR} \(\uparrow\) & \textbf{SSIM} \(\uparrow\) & \textbf{PSNR} \(\uparrow\) & \textbf{SSIM} \(\uparrow\) & \textbf{PSNR} \(\uparrow\) & \textbf{SSIM} \(\uparrow\) & \textbf{PSNR} \(\uparrow\) & \textbf{SSIM} \(\uparrow\) & \textbf{PSNR} \(\uparrow\) & \textbf{SSIM} \(\uparrow\) \\ \midrule
WDiffusion~\citep{ozdenizci2023restoring} & \(31.74\) & \(0.8768\) & \(31.79\) & \(0.8784\) & \(31.92\) & \(0.8809\) & \(30.31\) & \(0.8784\) & \(30.02\) & \(0.8716\) & \(30.82\) & \(0.8746\) \\
TransWeather~\citep{valanarasu2022transweather} & \(31.11\) & \(0.8684\) & \(31.13\) & \(0.8699\) & \(31.26\) & \(0.8741\) & \(29.24\) & \(0.8662\) & \(28.95\) & \(0.8565\) & \(29.15\) & \(0.8632\) \\
AirNet~\citep{li2022all} & \(32.46\) & \(0.8873\) & \(32.46\) & \(0.8887\) & \(32.75\) & \(0.8929\) & \(30.71\) & \(0.8874\) & \(30.40\) & \(0.8806\) & \(31.16\) & \(0.8825\) \\
PromptIR~\citep{potlapalli2023promptir} & \(31.18\) & \(0.8843\) & \(32.19\) & \(0.8867\) & \(32.45\) & \(0.8900\) & \(30.79\) & \(0.8903\) & \(30.43\) & \(0.8821\) & \(31.19\) & \(0.8847\) \\
EDVR~\citep{wang2019edvr} & \(28.70\) & \(0.7224\) & \(28.37\) & \(0.6991\) & \(29.07\) & \(0.7289\) & \(26.75\) & \(0.7259\) & \(26.94\) & \(0.7382\) & \(28.71\) & \(0.7675\) \\
BasicVSR++~\citep{chan2022basicvsr++} & \(33.22\) & \(0.9204\) & \(33.07\) & \(0.9180\) & \(33.32\) & \(0.9210\) & \(30.90\) & \(0.9048\) & \(30.52\) & \(0.8965\) & \(31.35\) & \(0.9011\) \\
ShiftNet~\citep{li2023simple} & \(33.09\) & \(0.9096\) & \(33.10\) & \(0.9113\) & \(33.34\) & \(0.9133\) & \(31.15\) & \(0.9027\) & \(30.82\) & \(0.8947\) & \(31.88\) & \(0.9000\) \\
RVRT~\citep{liang2022recurrent} & \(33.99\) & \(0.9314\) & \(33.98\) & \(0.9311\) & \(34.10\) & \(0.9315\) & \(31.73\) & \(0.9192\) & \(31.39\) & \(0.9113\) & \(32.47\) & \(0.9178\) \\
AverNet~\citep{zhaoavernet} & \(34.07\) & \(0.9333\) & \(34.09\) & \(0.9339\) & \(34.28\) & \(0.9356\) & \(31.73\) & \(0.9219\) & \(31.47\) & \(0.9145\) & \(32.45\) & \(0.9189\) \\ \midrule
\rowcolor{rowcolor} \textsc{\textbf{Ronin}} & \(33.68\) & \(\mathbf{0.9389}\) & \(33.82\) & \(\mathbf{0.9408}\) & \(33.84\) & \(0.9411\) & \(\mathbf{32.05}\) & \(\mathbf{0.9504}\) & \(\mathbf{32.11}\) & \(\mathbf{0.9510}\) & \(32.20\) & \(\mathbf{0.9523}\) \\ \bottomrule
\end{tabular}
}
\caption{\textbf{Time-Varying Unknown Degradation (TUD) Benchmark Results.} Quantitative results (PSNR and SSIM) on the TUD benchmark~\citep{zhaoavernet} comparing prior restoration methods.}
\label{tab:tud_denoise}
\end{table*}

\subsection{3D and 4D Benchmarks}

We consider two benchmarks for multi-degradation setting following the standard in all-in-one image restoration~\citep{li2022all,potlapalli2023promptir,conde2024high}, namely \(3\)D and \(4\)D. In \(3\)D benchmark, there are three different tasks: video deblurring, video denoising, and video deraining, while in \(4\)D benchmark, there are four different tasks: video deblurring, video denoising, video deraining and video desnowing. To not further segregate, we employ widely used standard restoration datasets for each task. To this end, for video deblurring, we use GoPro dataset~\citep{Nah_2017_CVPR}, and for video denoising we use DAVIS dataset~\citep{Pont-Tuset_arXiv_2017} adding white Gaussian noise with \(\sigma = 50\). Further, in the case of video deraining, we use VRDS dataset~\citep{wu2023mask}, and for video desnowing we employ RVSD~\citep{chensnow} which has both snow and haze degradations. Additional details on the datasets are presented in the appendix.

\begin{figure*}[!t]
    \centering
    \includegraphics[width=\linewidth]{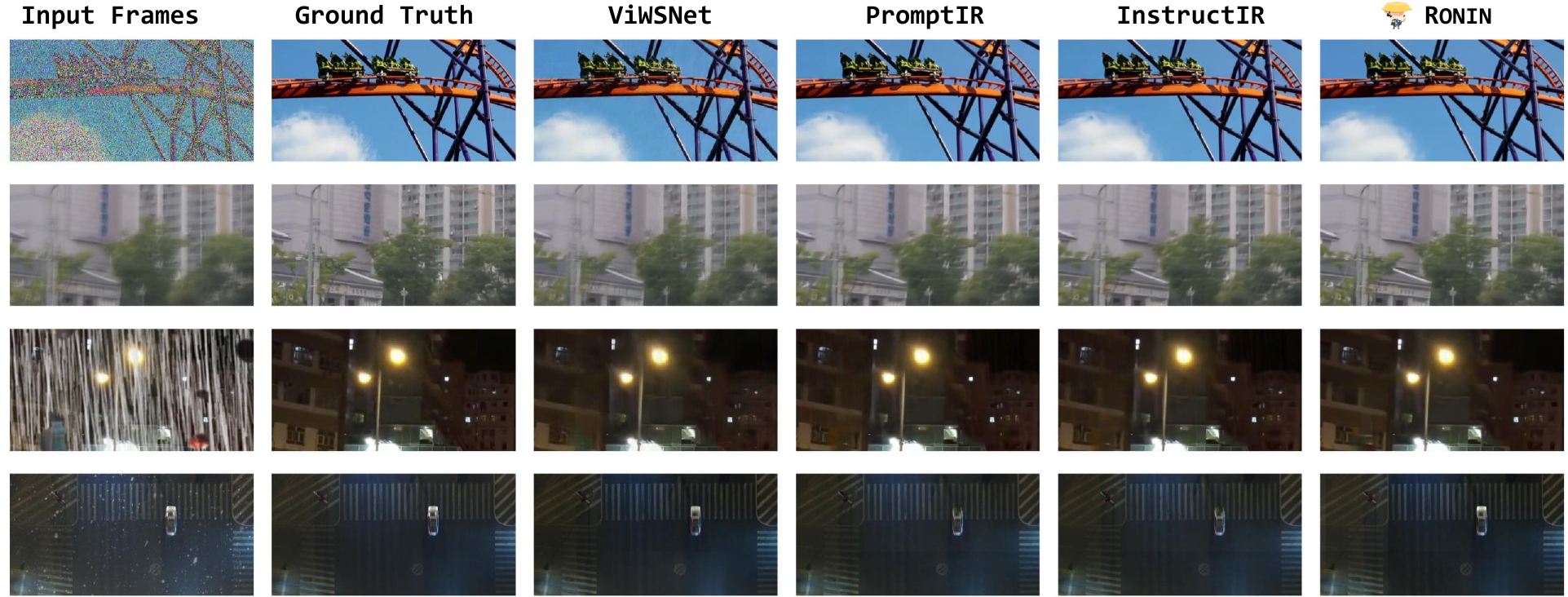}
    \caption{\textbf{Visual Results on \(\mathbf{4}\)D Benchmark.} We qualitatively compare three prior methods with \textsc{Ronin} on all tasks of the \(4\)D benchmark. The first row contains frame crops from denoising video, while the second, third, and fourth row contain frames crops from deblurring, deraining, and desnowing videos, respectively. Notice how \textsc{Ronin}'s outputs are faithful to the ground truth e.g., the cloud and vertical rollercoaster rods in the denoising video, trees and buildings in the deblurring video, light and the windows on buildings in the background in deraining video, and the car in the desnowing video. All of these regions in other methods' outputs show unwanted artifacts. Best viewed zoomed-in.}
    \label{fig:4dvisuals}
\end{figure*}

We present results on both benchmarks in~\cref{tab:3d_4d_results}. On \(3\)D benchmark, \textsc{Ronin} scores an average PSNR of \colorbox{rowcolor}{\(32.36\)} dB which is about \textcolor{goodgreen}{\(\mathbf{+1.15}\)} dB higher than the next best result. We also find that since AverNet~\citep{zhaoavernet} depends on optical flow for motion estimation, it noticeably suffers when the degradation is intense (e.g., in case of \(\sigma = 50\) noise). On \(4\)D benchmark, \textsc{Ronin} outperforms previous methods significantly by \textcolor{goodgreen}{\(\mathbf{+0.64}\)} dB. We also present visual results in~\cref{fig:3dvisuals} (for \(3\)D) and in~\cref{fig:4dvisuals} (for \(4\)D), and show that \textsc{Ronin} recovers videos such that they are more faithful to the ground truth and visually pleasing to the human eye.

\subsection{Time-Varying Unknown Degradations}
In~\citep{zhaoavernet}, the authors propose that degradations in videos can vary with time, and propose a time-varying unknown degradations benchmark where noise, blur, and compression intensity vary.\footnote{The noise is Gaussian (\(\sigma \in \mathcal{U}[10,15]\)), Speckle, and Poisson noise, while the blur is Gaussian and resize. The compression simulates different codecs for videos or the JPEG compression is simulated.} Following the said work, we use the dataset synthesized by the authors to train and test \textsc{Ronin}. We report the results in~\cref{tab:tud_denoise} on two datasets DAVIS~\citep{Pont-Tuset_arXiv_2017} and Set8~\citep{tassano2019dvdnet} and three settings \(t \in [6, 12, 24]\), i.e., degradation changes every \(6, 12\) or \(24\) frames, respectively. On the Set8 dataset, which has much longer videos than DAVIS, \textsc{Ronin} outperforms the previous best method AverNet~\citep{zhaoavernet} by an average of \textcolor{goodgreen}{\(\mathbf{+0.23}\)} dB on PSNR. While on the DAVIS testset, \textsc{Ronin} stays comparable on PSNR but outperforms prior art on the SSIM metric. We provide the qualitative analysis on all three settings in the appendix.

\begin{figure}
    \centering
    \includegraphics[width=\linewidth]{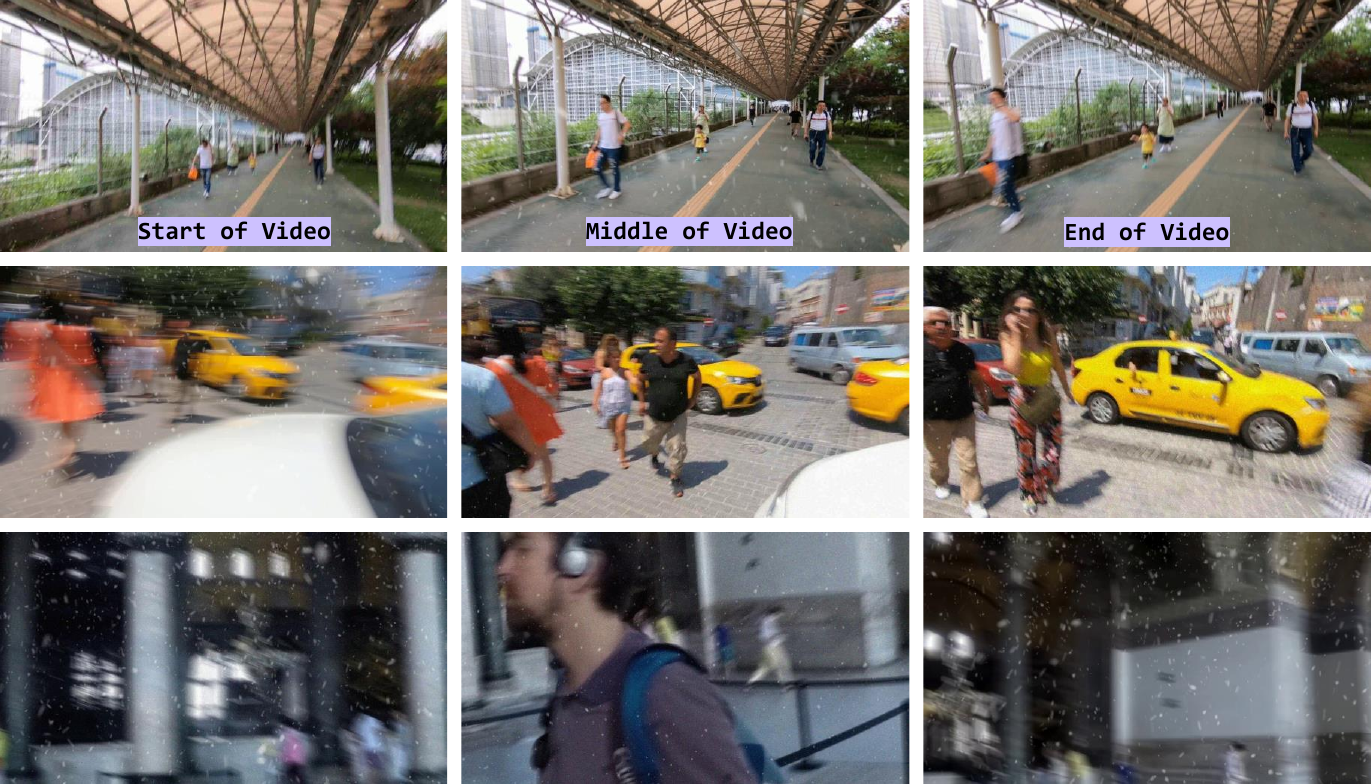}
    \caption{\textbf{Samples from the \texttt{SnowyScenes} Benchmark}. We present three frames from three different videos in the proposed \texttt{SnowyScenes} dataset. The first column includes frames sampled from early in the video, while the second and third columns include frames from the middle and end of the video, respectively.}
    \label{fig:snowyscenes}
\end{figure}

\subsection{Time-Varying Snow Degradations}
\begin{figure*}
    \centering
    \includegraphics[width=\linewidth]{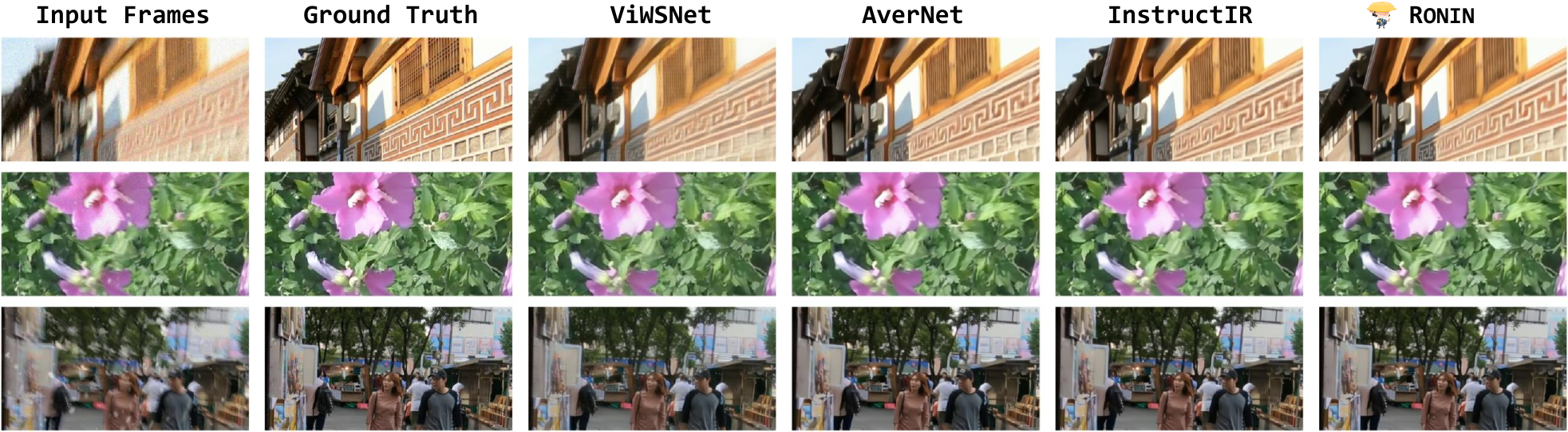}
    \caption{\textbf{Visual Results on \texttt{SnowyScenes} Benchmark.} We qualitatively compare three prior methods with \textsc{Ronin} on all tasks of the \(4\)D benchmark. The first row contains frame crops from \(t=6\) video, while the second and third rows contain frame crops from \(t=12\) and \(t=24\) videos, respectively. \textsc{Ronin}'s outputs are visually pleasing e.g., consider the pattern on the tile and grill on the window in the first video, the leaves in the second, and the photo-frames placed in the back and face of the person in grey shirt in the third video. All of these regions in other methods' outputs show unwanted artifacts. Best viewed zoomed-in.}
    \label{fig:snowyscenes_visuals}
\end{figure*}

We introduce another time-varying benchmark extending the TUD dataset proposed in~\citep{zhaoavernet}, termed as \texttt{SnowyScenes}. In TUD, synthetic noise, blur and compression artifacts are added to create three different variation settings (\(t \in [6, 12, 24]\)). In our proposed \texttt{SnowyScenes}, only noise and compression are synthetically synthesized following the TUD benchmark. However, unlike TUD, which uses Gaussian or resize blur, \texttt{SnowyScenes} builds on widely used video deblurring datasets, GoPro~\citep{Nah_2017_CVPR} and REDS~\citep{Nah_2019_CVPR_Workshops_REDS}, where the blur is realistic and scenes are dynamic. Further, we synthesize two sets of the same videos with different snow intensities, moderate and severe snow, using DaVinci Resolve\footnote{\href{https://www.blackmagicdesign.com/products/davinciresolve}{https://www.blackmagicdesign.com/products/davinciresolve}} and Python. We also include Poisson noise, Gaussian noise, speckle noise, video compression and JPEG compression artifacts, and follow the same procedure as outlined in~\citep{zhaoavernet} to create a corrupted video containing time-varying snow degradations, see~\cref{fig:snowyscenes} for a few frame samples. \texttt{SnowyScenes} contains \(20\) random videos from REDS train set and \(22\) random videos from the GoPro train set for a total of \(42\) training videos, and \(14\) test set videos with \(3\) from REDS test set and rest from GoPro test set. For the train set, the interval \(t\) of variation is set to \(6\), while for test set we consider three different intervals, i.e., \(6, 12, 24\), following~\citep{zhaoavernet}, to get three test sets. In other words, the degradations, including snow intensity, varies every \(t\) frames in the video. We report results and compare \textsc{Ronin} with representative all-in-one image and video restoration methods in~\cref{tab:snowyscenes_results}. On average, \textsc{Ronin} scores \textcolor{goodgreen}{\(\mathbf{+0.48}\)} dB higher than prior methods, and outperforms on all three settings. Further, we present visual results in~\cref{fig:snowyscenes_visuals}, and it can be seen that \textsc{Ronin} recovers the videos that are more faithful to the ground truth and visually pleasing to eye.

\begin{table}[]
\centering
\scalebox{0.83}{
\begin{tabular}{@{}lcccccc@{}}
\toprule
\multirow{2}{*}{\textbf{Method}} & \multicolumn{2}{c}{\(\mathbf{t = 6}\)} & \multicolumn{2}{c}{\(\mathbf{t = 12}\)} & \multicolumn{2}{c}{\(\mathbf{t = 24}\)} \\ \cmidrule(l){2-7} 
 & \textbf{PSNR} & \textbf{SSIM} & \textbf{PSNR} & \textbf{SSIM} & \textbf{PSNR} & \textbf{SSIM} \\ \midrule
AirNet~\citep{li2022all} & \(23.41\) & \(0.62\) & \(23.51\) & \(0.64\) & \(23.44\) & \(0.61\) \\
AverNet~\citep{zhaoavernet} & \(22.34\) & \(0.58\) & \(21.93\) & \(0.58\) & \(21.88\) & \(0.55\) \\
InstructIR~\citep{conde2024high} & \(29.56\) & \(0.91\) & \(29.63\) & \(0.91\)  & \(29.66\) & \(0.91\)  \\
PromptIR~\citep{potlapalli2023promptir} & \(29.72\) & \(0.91\) & \(29.79\) & \(0.91\) & \(29.81\) & \(0.91\) \\
ViWSNet~\citep{yang2023video} & \(27.22\) & \(0.87\) & \(27.27\) & \(0.87\) & \(27.33\) & \(0.87\) \\ \midrule
\rowcolor{rowcolor} \textsc{\textbf{Ronin}} & \(\mathbf{30.21}\) & \(\mathbf{0.92}\) & \(\mathbf{30.28}\) & \(\mathbf{0.92}\) & \(\mathbf{30.27}\) &  \(\mathbf{0.92}\) \\ \bottomrule
\end{tabular}
}
\caption{\textbf{\texttt{SnowyScenes} Benchmark Results.} Quantitative results (PSNR and SSIM) on the \texttt{SnowyScenes} benchmark comparing all-in-one restoration prior methods.}
\label{tab:snowyscenes_results}
\end{table}

\subsection{Discussion}
It is desirable to leverage the benefits of whitebox~\citep{conde2024high} and blackbox prompt~\citep{potlapalli2023promptir,zhaoavernet} methods. In whitebox prompt methods, interpretability is retained at the cost of a tightly coupled text encoder or MLLM/vision-language model (VLM) during inference. On the flip side, the blackbox prompts offer a standalone procedure to condition the restoration method in all-in-one paradigm. \textsc{Ronin} combines the best of both worlds by injecting whitebox prompts grounded in language but ensures that restoration functions standalone without relying on any text encoder or MLLM in inference. We observe that existing methods such as InstructIR~\citep{conde2024high}, AirNet~\citep{li2022all}, and ViWSNet~\citep{yang2023video} suffer with composite degradations, as seen in TUD and \texttt{SnowyScenes} benchmarks. These methods rely on class-level information which assumes that only a single degradation can corrupt the image (or video). The human-aligned instructions used by InstructIR~\citep{conde2024high} are also tailored to one degradation per input and are rigid by design. AverNet~\citep{zhaoavernet} injects blackbox prompts in the restoration backbone, but relies on optical flow to compensate for motion guided by these blackbox prompts. Consequently, other than the lack of interpretability, it also suffers in the presence of severe degradation (see denoising results in~\cref{tab:3d_4d_results} where \(\sigma = 50\)). Further, AverNet requires frames both in the future and in history to function due to the bidirectional propagation mechanism, and hence, is likely to suffer in the case of streaming videos. However, \textsc{Ronin} operates on a frame-by-frame basis and is capable of supporting online video restoration setting where frames arrive in sequential order.

% Few pointers to explain here
% \begin{itemize}
%     \item \textsc{Ronin} is prior free and does not require any external module in inference.
%     \item \textsc{Ronin}'s design allows it to restore composite degradations such as TUD and \textsc{SnowyScenes}. Methods like InstructIR and AirNet or ViWSNet which rely on a classification loss suffer.
%     \item \textsc{Ronin} functions frame-by-frame and can work for online video restoration setting where frames arrive in sequential ordering.
% \end{itemize}
\section{Ablation Study}
\label{sec:ablation}

\begin{table}[!t]
\centering
\scalebox{0.80}{
\begin{tabular}{@{}lcccccc@{}}
\toprule
\multirow{2}{*}{\textbf{\begin{tabular}[c]{@{}c@{}}Prompt\\ Location\end{tabular}}} & \multicolumn{2}{c}{\begin{tabular}[c]{@{}c@{}}\textbf{Deblur}\\ (GoPro~\citep{Nah_2017_CVPR})\end{tabular}} & \multicolumn{2}{c}{\begin{tabular}[c]{@{}c@{}}\textbf{Denoise}\\ (DAVIS~\citep{Pont-Tuset_arXiv_2017})\end{tabular}} & \multicolumn{2}{c}{\begin{tabular}[c]{@{}c@{}}\textbf{Derain}\\ (VRDS~\citep{wu2023mask})\end{tabular}} \\ \cmidrule(l){2-7} 
 & \textbf{PSNR} & \textbf{SSIM} &  \textbf{PSNR} & \textbf{SSIM} &  \textbf{PSNR} & \textbf{SSIM} \\ \midrule
 \begin{tabular}[c]{@{}c@{}}Turtle~\citep{ghasemabadilearning} \end{tabular} & \(28.68\) & \(0.914\) & \(30.59\) & \(0.906\) & \(29.01\) & \(0.934\) \\
\begin{tabular}[c]{@{}c@{}}No Prompt \end{tabular} & \(28.81\) & \(0.915\) & \(30.48\) & \(0.906\) & \(29.03\) & \(0.925\) \\
First Decoder & \(28.89\) & \(0.916\) & \(30.56\) & \(0.906\) & \(29.11\) & \(0.930\) \\
All Decoders & \(28.97\) & \(0.918\) & \(30.64\) & \(0.908\) & \(29.19\) & \(0.938\) \\ \midrule
\rowcolor{rowcolor} \textsc{\textbf{Ronin}} & \(\mathbf{28.99}\) & \(\mathbf{0.919}\) & \(\mathbf{30.65}\) & \(\mathbf{0.908}\) & \(29.18\) & \(0.937\) \\ \bottomrule
\end{tabular}
}
\caption{\textbf{Ablating Prompt Placement.} Results of ablating the prompt injection module and utility of \textsc{Ronin}'s design choices.}
\label{tab:ablation_prompt}
\end{table}

%We ablate prompt placement in \textsc{Ronin} architecture~\cref{tab:ablation_prompt} on the \(3\)D setting. All the models are similar in size in terms of number of parameters and MACs (G), but we reduce the available budget to \(3\)M parameters and adjust the model design accordingly. We consider three variations wherein the prompt is injected in either first decoder, all the decoders or last two decoders (\textsc{Ronin}) and find that the best results are obtained when prompt is injected in the last two decoders indicating that channel modulation based on language yields maximum benefit at those stages. We also ablate if prompt is necessary in the first place (No Prompt setting) and find that prompt considerably helps with the performance since it allows greater flexibility in how model adapts to different degradations. Further ablation studies are deferred to the appendix.

We ablate prompt placement in \textsc{Ronin} on the \(3\)D benchmark, see results in~\cref{tab:ablation_prompt}. All the models are similar in size in terms of the number of parameters and MACs (G), with a budget of \(3\)M parameters. We consider three variations wherein the prompt is injected in either the first decoder, all the decoders, or the last two decoders (\textsc{Ronin}). Our findings indicate that the best results are achieved when the prompt is injected into the last two decoders, suggesting that channel modulation based on language provides the greatest benefits at these stages. Further, we examine the necessity of using a prompt (No Prompt setting) and conclude that using a prompt significantly improves performance, as it offers greater flexibility for the model to adapt to various degradations. In the No Prompt setting in~\cref{tab:ablation_prompt}, the cross-attention between the first encoder and the latent stage to introduce degradation information back in is still present. We consider another setting where we compare \textsc{Ronin} to Turtle~\citep{ghasemabadilearning} on the \(3\)D benchmark. We observe that \textsc{Ronin} outperforms the base Turtle architecture, indicating the utility of injecting grounded knowledge. Note that additional ablation studies and discussions are deferred to the appendix~\cref{appdx:moreablation}.

% \begin{table}[]
% \centering
% \scalebox{0.85}{
% \begin{tabular}{@{}ccccccc@{}}
% \toprule
% \multirow{2}{*}{\begin{tabular}[c]{@{}c@{}}\textbf{Prompt} \\ \textbf{Style}\end{tabular}} & \multicolumn{2}{c}{\begin{tabular}[c]{@{}c@{}}\textbf{Deblur}\\ (GoPro~\citep{Nah_2017_CVPR})\end{tabular}} & \multicolumn{2}{c}{\begin{tabular}[c]{@{}c@{}}\textbf{Denoise}\\ (DAVIS~\citep{Pont-Tuset_arXiv_2017})\end{tabular}} & \multicolumn{2}{c}{\begin{tabular}[c]{@{}c@{}}\textbf{Derain}\\ (VRDS~\citep{wu2023mask})\end{tabular}} \\ \cmidrule(l){2-7} 
%  & \textbf{PSNR} & \textbf{SSIM} & \textbf{PSNR} & \textbf{SSIM} & \textbf{PSNR} & \textbf{SSIM} \\ \midrule
% \begin{tabular}[c]{@{}c@{}}Instructions \end{tabular} &  \(30.93\) & \(0.94\) & \(31.25\) & \(0.92\) & \(31.10\) & \(0.95\) \\
% \begin{tabular}[c]{@{}c@{}}Degradations \\ Grounded in \\ Language\\ (\textsc{Ronin})\end{tabular} &  &  &  &  &  &  \\ \bottomrule
% \end{tabular}
% }
% \caption{\textbf{Ablating the Style of Prompts.}}
% \label{tab:ablation_prompt_style}
% \end{table}

% \paragraph{Prompt Style}
\section{Conclusion}
\label{sec:conc}

We introduced \textsc{Ronin}, an all-in-one video restoration method that uses a multimodal large language model (MLLM) to ground degradations in natural language. \textsc{Ronin} learns to approximate the necessary information during training, allowing the MLLM to be safely removed during inference without any extra cost, but offering interpretable conditioning. We also introduced \texttt{SnowyScenes}, a dynamic snow intensity dataset, extending time-varying degradation to weather. By standardizing all-in-one video benchmarks, we hope that this work paves the way for future research in low-level vision, particularly for videos.

% Our method outperforms existing all-in-one image and video restoration methods on multiple benchmarks, including \(3\)D, \(4\)D, TUD, and \texttt{SnowyScenes}. 

% We also extend the time-varying degradation setting to weather by introducing a dynamic snow condition dataset, \texttt{SnowyScenes}. 
% In this work, we attempted to draw attention towards the all-in-one video restoration problem, which is more a general and natural setting than its image counterpart. We proposed an all-in-one video restoration method, \textsc{Ronin}, that leverages a vision-language model (VLM) to ground the degradations in natural language. Unlike prior methods, \textsc{Ronin} is not tightly coupled with the VLM since it learns to approximate the necessary information during training. As such, the VLM is safely removed during inference adding no extra cost to the underlying restoration method, but allowing interpretable conditioning to adapt a single model to restore videos afflicted by multiple degradations. We also extend the time-varying composite degradation setting to weather by introducing a dataset with dynamic snow conditions in videos. We compared \textsc{Ronin} with standard video and image all-in-one restoration methods in the literature and reported state-of-the-art performance on three-task (3D), four-task (4D), Time-Varying Degradation (TUD) and our proposed Time-Varying Snow Degradation (\texttt{SnowyScenes}) benchmarks. By standardizing all-in-one video benchmarks, we hope that this work paves the way for future research in low-level vision, particularly for videos.

{
    \small
    \bibliographystyle{ieeenat_fullname}
    \bibliography{main}
}
\clearpage
\setcounter{page}{1}
\twocolumn[{%
\renewcommand\twocolumn[1][]{#1}%
\maketitlesupplementary
\begin{center}
    \centering
    \captionsetup{type=figure}
    \includegraphics[width=\textwidth]{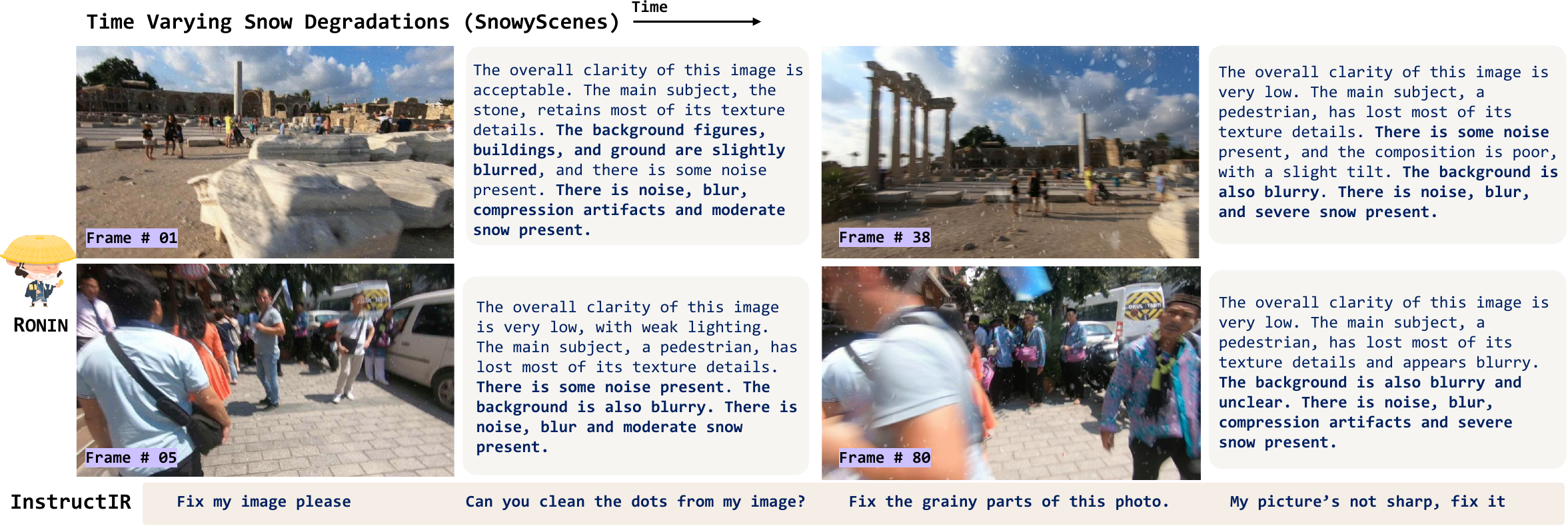}
    \captionof{figure}{\textbf{Differences in Grounding Degradations and Instructions.} We sample two frames (at different timesteps) from two different videos of \texttt{SnowyScenes} benchmark and compare \textsc{Ronin}'s language grounded descriptions with InstructIR~\citep{conde2024high}'s human-aligned instructions. Since InstructIR randomly samples instructions for each degradation, we show two samples (second and third) taken from noisy and blurry instructions, while first and third samples are taken from general instructions. It is evident that instructions are rigid and provide no meaningful clue without identifying the degradations. \textsc{Ronin} benefits from per-frame grounded degradations that also describe context.}
    \label{fig:instruct_rig}
\end{center}%
}]

\section*{Appendices}

\section{Additional Ablation Studies}
\label{appdx:moreablation}
We discuss the motivation behind grounding degradations, and present additional ablation studies to further understand different components of \textsc{Ronin} and the design choices made.

\subsection{Motivation: Grounding Degradations}
We posit that grounding the degradations in natural language to serve as a prior for the restoration algorithm offers flexible control along with interpretability. The instruction condition in methods such as InstructIR~\citep{conde2024high}, although interpretable, requires that for each input, a random degradation-specific instruction is sampled and fed as input to the restoration method. While this is plausible in images, videos are much more challenging. Consider how restoring a \(30\)fps \(10\) seconds video is dependent on \(300\) different calls to the text encoder in InstructIR~\citep{conde2024high}, the VLM in~\citep{luo2023controlling} or the MLLM in~\citep{jin2024llmra}. We ablate this limitation in InstructIR~\citep{conde2024high} where we consider a single instruction variation i.e., we sample a degradation-dependent instruction once and re-use it for all the videos in the same degradation category and report results on the \(3\)D benchmark in~\cref{tab:instructIR_variation}. Unsurprisingly, InstructIR~\citep{conde2024high} observes non-trivial performance drop.

In \textsc{Ronin}, however, no such limitation exists due to the proposed prompt approximation objective allowing the MLLM to be safely removed post-training. Further, grounded conditioning allows nuances in modulating channels since plain instructions can be rigid (e.g., \textit{`clean up this image'}) and cannot handle composite degradations without complete knowledge of degradations at the inference time. Since the natural language grounding in \textsc{Ronin} also captures the context of the frame and offers fine-grained control, our proposed method is a positive step towards designing region-specific restoration methods (e.g., the sky has high noise due to flat texture, the building has ghosting artifacts due to repetitive patterns, etc.). We illustrate this further in~\cref{fig:instruct_rig} where we show that instructions that InstructIR~\citep{conde2024high} leverages are indeed rigid and fail to capture composite degradations meaningfully.

\begin{table}[!t]
\centering
\scalebox{0.85}{
\begin{tabular}{@{}lcccccc@{}}
\toprule
\multicolumn{1}{c}{\multirow{2}{*}{\textbf{Method}}} & \multicolumn{2}{c}{\begin{tabular}[c]{@{}c@{}}\textbf{Deblur}\\ (GoPro~\citep{Nah_2017_CVPR})\end{tabular}} & \multicolumn{2}{c}{\begin{tabular}[c]{@{}c@{}}\textbf{Denoise}\\ (DAVIS~\citep{Pont-Tuset_arXiv_2017})\end{tabular}} & \multicolumn{2}{c}{\begin{tabular}[c]{@{}c@{}}\textbf{Derain}\\ (VRDS~\citep{wu2023mask})\end{tabular}} \\ \cmidrule(l){2-7} 
\multicolumn{1}{c}{} & \textbf{PSNR} & \textbf{SSIM} & \textbf{PSNR} & \textbf{SSIM} & \textbf{PSNR} & \textbf{SSIM} \\ \midrule
InstructIR~\citep{conde2024high} & \multicolumn{1}{l}{\(30.93\)} & \multicolumn{1}{l}{\(0.94\)} & \multicolumn{1}{l}{\(31.25\)} & \multicolumn{1}{l}{\(0.92\)} & \multicolumn{1}{l}{\(31.10\)} & \multicolumn{1}{l}{\(0.95\)} \\
\begin{tabular}[c]{@{}l@{}}Single \\ Instruction\end{tabular} & \(\mathbf{\color{red}30.90}\) & \(0.94\) & \(\mathbf{\color{red}31.17}\) & \(0.91\) & \(\mathbf{\color{red}31.03}\) & \(0.95\) \\ \midrule
\rowcolor{rowcolor} \textsc{\textbf{Ronin}}  & \(\textbf{32.73}\) & \(\textbf{0.96}\) & \(\textbf{31.65}\) & \(\textbf{0.92}\)  & \(\textbf{32.72}\) &  \(\textbf{0.97}\) \\ \bottomrule
\end{tabular}
}
\caption{\textbf{Frequency of Instruction Sampling.} Results on \(3\)D benchmark ablating the frequency of instruction sampling in InstructIR~\citep{conde2024high}, and comparison with \textsc{Ronin} which does not need any instructions/text during inference.}
\label{tab:instructIR_variation}
\end{table}

\begin{figure}
    \centering
    \includegraphics[width=\linewidth]{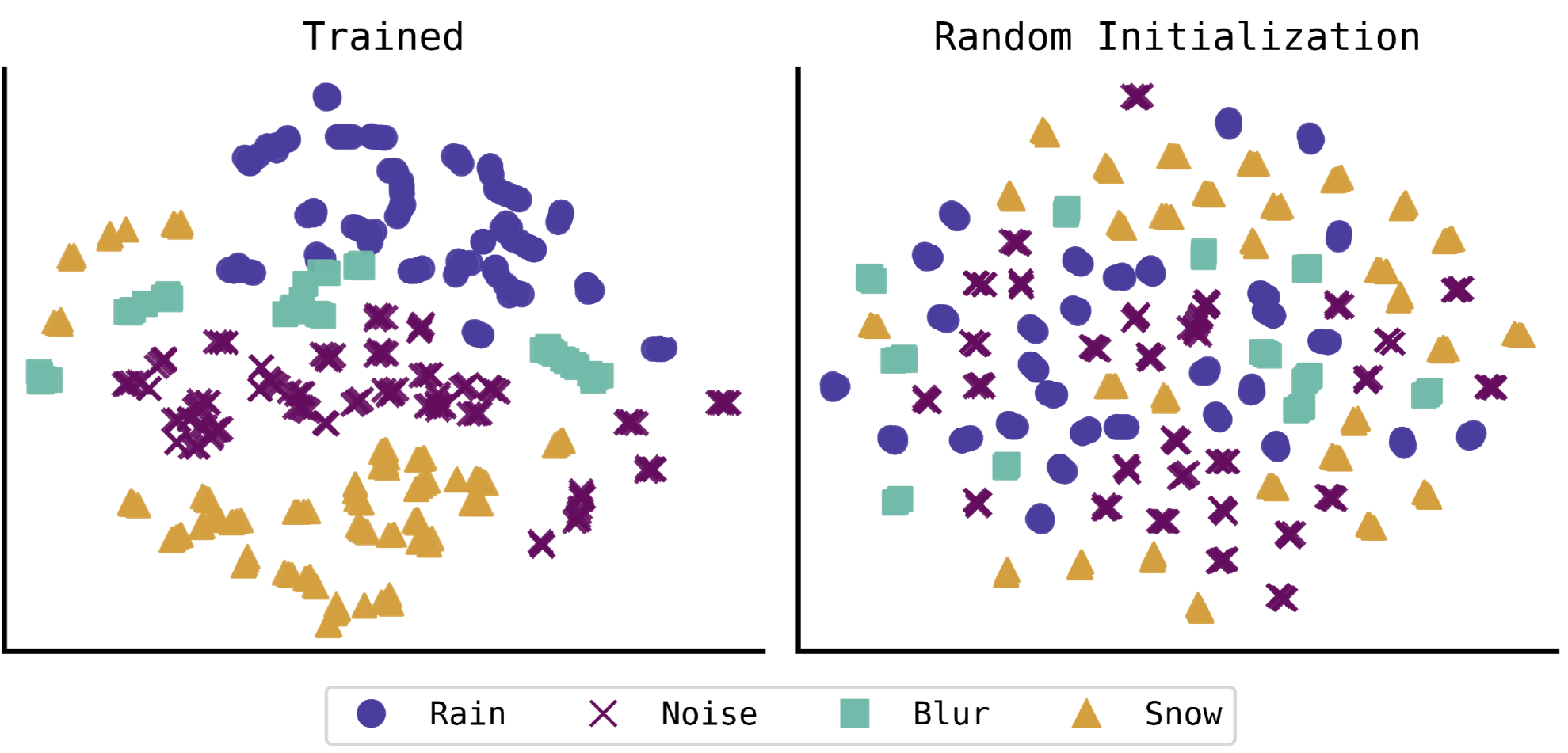}
    \caption{\textbf{tSNE Plot.} Visualization of learned and untrained prompts taken from the latent space of \textsc{Ronin} on \(4\)D benchmark.}
    \label{fig:tSNEplot}
\end{figure}

\paragraph{Are Learned Prompts Meaningful?} To illustrate that the learned prompts are meaningful, we perturb the learned prompts with white Gaussian noise in inference and evaluate on \(3\)D benchmark, see~\cref{tab:perturbed_prompts}. We observe a significant drop in performance indicating that if wrong prompt information were propagated, \textsc{Ronin} would suffer. The drop in the performance illustrates that the learned prompts modulate the output and are necessary for the observed performance gains. We also visualize tSNE plots of learned and untrained prompts, showing that learned prompts effectively differentiate between degradations, see~\cref{fig:tSNEplot}. Further, we also compute cosine similarity between the learned prompts and the raw text embedding taken from the text encoder and compare it with random prompts (untrained). We find that in the former case, trained prompts align closely with raw text embeddings (similarity scores in range of 0.9852–0.9914), while random prompts do not (similarity scores in range of -0.0393–0.0370). 

% \paragraph{Is Language Prompt Necessary?} 

\begin{table}[!t]
\centering
\scalebox{0.8}{
\begin{tabular}{@{}lccccc@{}}
\toprule
\textbf{Methods} & \textbf{Denoise} & \textbf{Deblur} & \textbf{Derain} & \textbf{MACs (G)} & \textbf{Params} \\ \midrule
InstructIR~\citep{conde2024high} & 0.1799 & 0.1444 & 0.0623 & \(133.73^{\textcolor{blue}{*}}\) & \(73.9^{\textcolor{blue}{*}}\)M \\
PromptIR~\citep{potlapalli2023promptir} & 0.1793 & 0.1293 & 0.0578 & \(158.49\) & \(35.6\)M \\
ViWSNet~\citep{yang2023video} & 0.1734 & 0.1890 & 0.0902 & \(88.93\) & \(57.7\)M \\
AverNet~\citep{zhaoavernet} & 0.4277 & 0.1394 & 0.0640 & \(127.72^{\textcolor{red}{*}}\) & \(41.3^{\textcolor{red}{*}}\)M \\ \midrule
\rowcolor{rowcolor} \textsc{Ronin} & \textbf{0.1713} & \textbf{0.1037} & \textbf{0.0463} & \(167.23\) & \(57\)M \\  \bottomrule
\vspace{-2em}
\end{tabular}
}
\caption{\textbf{Perceptual Results.} LPIPS scores on \(3\)D benchmark (\(\downarrow\) is better), with MACs (G) and number of parameters Params (M). \(^{\textcolor{red}{*}}\) indicates that optical flow network, while \(^{\textcolor{blue}{*}}\) indicates that the text encoder parameters were not included.}
\label{tab:lpips}
\end{table}

\paragraph{Perceptual Results of \textsc{Ronin}} On \(3\)D benchmark, we present LPIPS~\citep{zhang2018unreasonable} scores and compare it to prior methods. In line with the qualitative results, \textsc{Ronin} scores better on the metric (lower is better) indicating that the restored videos are pleasing to the human eye.

\section{Additional Related Work}
\label{appdx:morerelatedwork}

Video restoration, in literature, is studied from several facets, mostly distributed in terms of how the motion is estimated and compensated for, and how the frames are processed in the learning procedure. Several methods employ optical flow to explicitly estimate motion, and devise a compensation strategy as part of the learning procedure, such as deformable convolutions~\citep{liang2022recurrent,liang2024vrt}, or flow refinement~\citep{huang2022neural}. On the other end, methods rely on the implicit learning of correspondences in the latent space across the temporal resolution of the video, a few strategies include temporal shift modules~\citep{li2023simple}, or non-local search~\citep{vaksman2021patch,li2020mucan,zhou2023exploring}. Further, similar differentiation exists in the manner a video is processed i.e., several methods opt for either recurrence in design~\citep{zhu2022deep,zhao2021recursive,zhong2020efficient} while others restore several frames at once~\citep{chen2021multiframe,haris2019recurrent}.

\begin{table}[!t]
\centering
\scalebox{0.85}{
\begin{tabular}{@{}lcccccc@{}}
\toprule
\multirow{2}{*}{\textbf{\begin{tabular}[c]{@{}c@{}}Prompt\\ Style\end{tabular}}} & \multicolumn{2}{c}{\begin{tabular}[c]{@{}c@{}}\textbf{Deblur}\\ (GoPro~\citep{Nah_2017_CVPR})\end{tabular}} & \multicolumn{2}{c}{\begin{tabular}[c]{@{}c@{}}\textbf{Denoise}\\ (DAVIS~\citep{Pont-Tuset_arXiv_2017})\end{tabular}} & \multicolumn{2}{c}{\begin{tabular}[c]{@{}c@{}}\textbf{Derain}\\ (VRDS~\citep{wu2023mask})\end{tabular}} \\ \cmidrule(l){2-7} 
 & \textbf{PSNR} & \textbf{SSIM} &  \textbf{PSNR} & \textbf{SSIM} &  \textbf{PSNR} & \textbf{SSIM} \\ \midrule
\begin{tabular}[c]{@{}c@{}}Perturbed\\ Prompts\end{tabular} & \(15.93\) & \(0.56\) & \(16.02\) & \(0.57\) & \(16.97\) & \(0.55\) \\ \midrule 
\rowcolor{rowcolor} \textsc{\textbf{Ronin}}  & \(\textbf{32.73}\) & \(\textbf{0.96}\) & \(\textbf{31.65}\) & \(\textbf{0.92}\)  & \(\textbf{32.72}\) &  \(\textbf{0.97}\) \\ \bottomrule
\end{tabular}
}
\caption{\textbf{Prompt Importance.} We perturb the prompts with white Gaussian noise and compute scores on the \(3\)D benchmark dataset. The significant drop in performance illustrates that the learned prompts modulate the output and are necessary for the observed performance gains.}
\label{tab:perturbed_prompts}
\end{table}

\subsection{All-In-One Image Restoration}
% Image and video restoration is well-studied in the literature~\citep{chen2022simple,zamir2022restormer,liang2021swinir,ghasemabadi2024cascadedgaze,liang2022recurrent,liang2024vrt,huang2022neural,li2023simple}. However, recently, there has been a surge in methods learning a single parameterized model to restore several degradations at once. 

There have been several methods introduced in the literature for the purpose of all-in-one image restoration. All of these methods utilize backbone architectures which are constructed in either columnar~\citep{liang2021swinir} or UNet~\citep{ronneberger2015u} fashion. Its extension to all-in-one tasks is aided by some conditioning on the restoration procedure, either only in the decoder (reconstruction), or conditioning at the latent stage. This condition is often realized in the form of some prior, either through degradation-aware feature injection, or through implicit (blackbox) or explicit (whitebox) prompts. However, all of the methods can be categorized into three different settings: contrastively learning the degradation information before restoring the input, implicitly injecting prompts to condition the restoration, or explicitly injecting prompts realized through degradation or textual features.

To the best of our knowledge, AirNet~\citep{li2022all} proposed the first standardized baseline all-in-one method to recover images from a variety of degradation levels and corruptions. The authors proposed a contrastive learning based degradation encoder that learned to differentiate between the degradations in its latent space. The following architecture then learned to restore the frames conditioned on the contrastively learned representations of the degraded input. Another all-in-one restoration method for weather-specific degradations is TransWeather~\citep{valanarasu2022transweather}. TransWeather proposed a Transformer-In-Transformer~\citep{han2021transformer} style encoder to learn hierarchical features, followed by a weather degradation queries conditioned decoder to recover the clean image. In both of these methods, some degradation-specific guidance is provided--class labels for positive and negative sample mining in the case of AirNet, or weather-specific queries in TransWeather. 

However, different from these, PromptIR~\citep{potlapalli2023promptir} proposed to inject prompts in the decoder of the encoder-decoder style restoration architecture. The prompts were implicitly learned since they were input-conditioned, and the method required no supervision on the degradation. Henceforth, a series of all-in-one image restoration methods followed the baselines set by AirNet~\citep{li2022all}, and proposed different architectures for the task. However, most of these works differ in how the degradation information is injected in the learning procedure, either implicitly or otherwise. Prompt-In-Prompt (PIP)~\citep{li2023prompt} proposed to fuse two prompts, i.e., degradation-aware prompt, and base restoration prompt, into a universal prompt. The resultant universal prompt is then fused with the input features through a feature-prompt mixing module for the restoration tasks.

Contemporary works such as InstructIR~\citep{conde2024high} proposed to inject human-aligned instructions into the restoration architecture's decoders through a prompt-feature mixing module. In practice, the instructions, generated through a multi-modal large language model, were first fed into a sentence transformer (pretrained on large textual data) to compute the instruction embeddings for the restoration procedure. One downside of such an approach is that on deployment, the sentence transformer can not be decoupled from the restoration architecture since the decoder is conditioned on the instruction embeddings obtained from the sentence transformer. Similarly, LLMRA~\citep{jin2024llmra} leveraged a multi-modal large language model (MLLM) to generate context descriptions, and a CLIP text encoder~\citep{radford2021learning} to obtain embeddings of the context. These embeddings were then injected into the restoration procedure. LLMRA suffers from similar limitations as InstructIR i.e., both of these methods have to deploy the underlying procedure used to generate embeddings along with the restoration architecture. In line with language-guided restoration, several methods such as LanguageWeather~\citep{yang2023language}, and TextIR~\citep{yan2023textual} also leverage language models (or vision-language models) to introduce degradation prior in the restoration procedure. 

\begin{figure*}[!t]
    \centering
    \includegraphics[width=\linewidth]{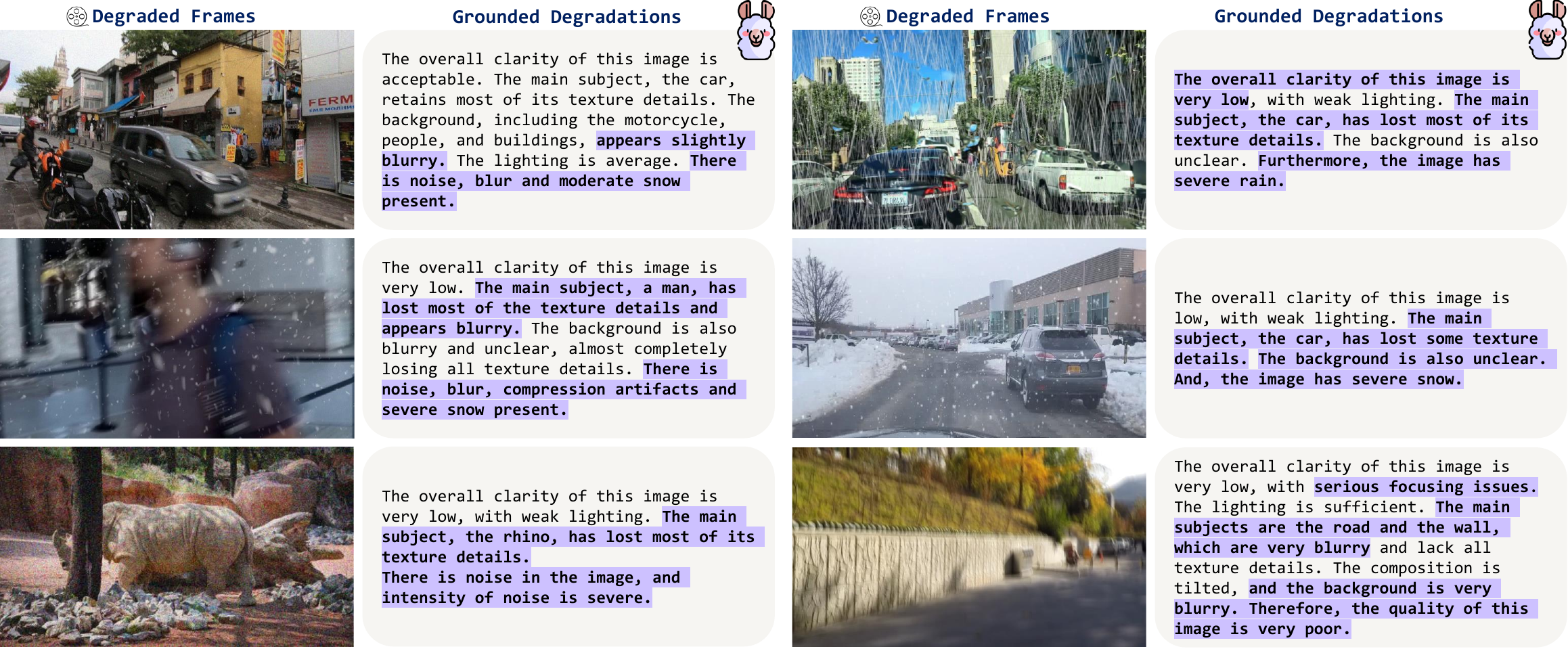}
    \caption{\textbf{Samples of Degradations Descriptions.} A few samples of frames and their respective grounded degradation prompts taken from different benchmarks. In the first column, from top to bottom, the frames are taken from \texttt{SnowyScenes} (moderate snow), \texttt{SnowyScenes} (severe snow), \(3\)D (denoise). In the second column, from top to bottom, the frames are taken from \(3\)D (derain), \(4\)D (desnow), and \(3\)D (deblur) benchmarks, respectively.}
    \label{fig:more_prompt_samples}
\end{figure*}

\subsection{All-In-One Video Restoration}
All of the image methods discussed above are comparable to each other given consistent evaluation on similar all-in-one restoration datasets and tasks. However, the all-in-one video restoration progress is siloed, and the attempts made in literature are disparate in nature. VJT~\citep{hui2024vjt} proposed a multi-degradation restoration architecture for low-light enhancement, deblurring and denoising tasks. The proposed Transformer-based architecture employed a multi-tier setup wherein each tier utilized a different level of degraded video as a target for feature learning process. Further, they also introduced a new Multi-scenes Lowlight-Blur-Noise (MLBN) dataset for the restoration task. However, the dataset was not publicly released for any follow-up methods to train and evaluate their methods on. Similarly, another work~\citep{shekarforoush2023dual} introduced joined deblurring and denoising method, and proposed a new dataset for the task. The proposed method departed from conventional architecture design in all-in-one restoration literature by introducing separate encoders for each task. However, similar to VJT, the dataset was not publicly released. Before VJT, another method CDUN~\citep{cheng2023cross} proposed an all-in-one video restoration architecture targeting deraining, dehazing, desnowing and low-light enhancement tasks. Although similar in a few tasks to VJT~\citep{hui2024vjt}, CDUN utilized different datasets, while synthesizing own video desnowing dataset due to, then, a lack of any video desnowing dataset. More recently, ViWS-Net~\citep{yang2023video} proposed all-in-one video restoration architecture for weather degradation removal, namely for desnowing, dehazing and deraining tasks. However, since CDUN~\citep{cheng2023cross} did not publicly release the desnowing dataset that they reported scores on, ViWS-Net synthesized another desnowing dataset, referred to as KITTI-Snow based on the KITTI dataset~\citep{liao2022kitti}\footnote{\url{https://github.com/scott-yjyang/ViWS-Net} KITTI-Snow was publicly released.}. More recently, AverNet~\citep{zhaoavernet} proposed time-varying degradation dataset where every fixed interval (a predefined frame, e.g., every sixth frame), the degradation changed simulating varying corruption in a video. The authors argue that this setting is more natural to videos. However, the degradations considered are limited to variations in noise, Gaussian blur and compression. 

\section{Dataset Details}
\label{appdx:datasetdetails}

All of the benchmarks considered in this work are created through standard datasets in video restoration literature and are available open-source for academic research purposes, except our proposed \texttt{SnowyScenes} benchmark, which will be open-sourced and released publicly for future research work.

\subsection{3D Benchmark}
As discussed earlier, we consider three different video restoration tasks to form the \(3\)D benchmark, namely video denoising, video deraining, and video deblurring. In video denoising, following~\citep{tassano2020fastdvdnet}, we employ the DAVIS~\citep{Pont-Tuset_arXiv_2017} dataset which consists of \(60\) videos in the training set and \(30\) videos in the held-out test set. We add white Gaussian noise with \(\sigma \in \mathcal{U}[20,50]\), and test with \(\sigma = 50\) Gaussian noise. In video deraining, we use the video raindrop and rain streak removal (VRDS) dataset introduced in~\citep{wu2023mask}. The dataset comprises videos captured in diverse scenarios in both daytime and nighttime settings corrupted by both rain streaks and raindrops. There are a total of \(102\) videos at a resolution of \(1280\times 720\) with \(100\) frames per video in the dataset, and \(72\) are in training set while \(30\) are in the held-out test set. In video deblurring, we employ the GoPro dataset introduced in~\citep{Nah_2017_CVPR} which contains videos captured from the GOPRO4 Hero consumer camera at a resolution of \(1280\times 720\). The dataset contains \(3214\) pairs of blurry and sharp images, with \(2103\) pairs in the training set and \(1111\) pairs in the test set. GoPro dataset is formed by integrating sharp information over time for blur image generation, instead of modeling a kernel to convolve on the sharp image~\citep{Nah_2017_CVPR}.

\begin{figure*}[!t]
    \centering
    \includegraphics[width=\linewidth]{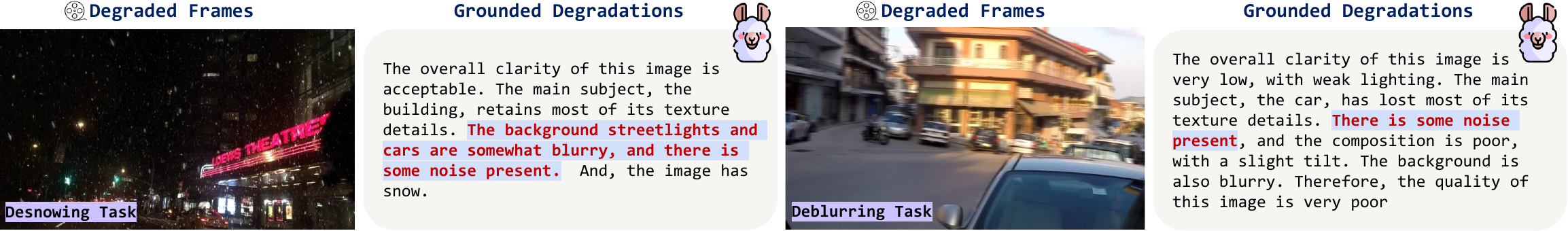}
    \caption{\textbf{Illustration of Limitation in Grounded Degradations.} Two samples of language descriptions where extraneous degradations are present. The first frame is taken from a desnowing task video, but the prompt describes \textit{noise and blur}. Although the frame has slight blur and arguably even noise, the ground truth is only free of snow degradation. The second frame is taken from a deblurring video, but there is mention of \textit{some noise} in the description.}
    \label{fig:limitations}
\end{figure*}

\subsection{4D Benchmark}
The \(4\)D benchmark considers four different video restoration tasks, with three being similar to the ones in \(3\)D benchmark. The additional restoration task is video desnowing and dehazing. In~\citep{chensnow}, the authors introduced a video desnowing and dehazing dataset, RVSD. The dataset consists of \(110\) videos at varying resolutions from \(480\)p to \(4\)k, with \(80\) videos in the training set and \(30\) videos in the held-out test set. RVSD contains dynamic scenes in varied lighting conditions, both in night and daytime, and has realistic and dynamic snow and haze rendered in Unreal Engine.

\subsection{SnowyScenes Benchmark}
In both \(3\)D and \(4\)D benchmarks, a single degradation affects a video, i.e., there are no videos with composite degradations. However, in many cases, degradations affect videos in a time-varying fashion. In other words, degradations change in intensity or even type as more frames are sampled/observed. To simulate such a setting, a new dataset called time-varying degradations, TUD, was introduced in a recent work~\citep{zhaoavernet}. In TUD, the authors considered degradations introduced by Gaussian, Poisson and Speckle noise, kernel-based blur, and video/JPEG compression. In this work, we propose a harder time-varying setting, \texttt{SnowyScenes}, with realistic blur and varying snow intensity. We pick \(56\) blurry videos from widely used GoPro~\citep{Nah_2017_CVPR} and REDS~\citep{Nah_2019_CVPR_Workshops_REDS} datasets, with \(42\) videos in the training set and \(14\) in the held-out test set. We borrow Gaussian, Poisson and Speckle noise and compression degradations, but synthesize snow with two intensity levels moderate and severe. For Gaussian and Speckle noise, the noise levels are sampled uniformly from \([10, 15]\), while the Poisson noise \(\alpha\) is sampled from \([2,4]\) following the Poisson noise mathematical model \(\mathcal{P}(10^{\alpha} \times x)/10^{\alpha} - x\). Further, in the case of compression, the quality factor in JPEG compression is randomly chosen from \(\{20, 30, 40\}\), while in video compression the codecs are randomly chosen from \(\{\text{libx264}, \text{h264}, \text{mpeg4}\}\), following~\citep{zhaoavernet}. Since the videos already have dynamic blur which is kernel-free, we do not further add Gaussian or resize blur. To generate a corrupted video, degradations are sampled with a probability of \(0.55\). We summarize the statistics of our proposed benchmark in~\cref{tab:snowyscenes_stats}. The benchmark will be released along with the necessary codebase for reproducibility and future research.

\begin{figure*}[!t]
    \centering
    \includegraphics[width=\linewidth]{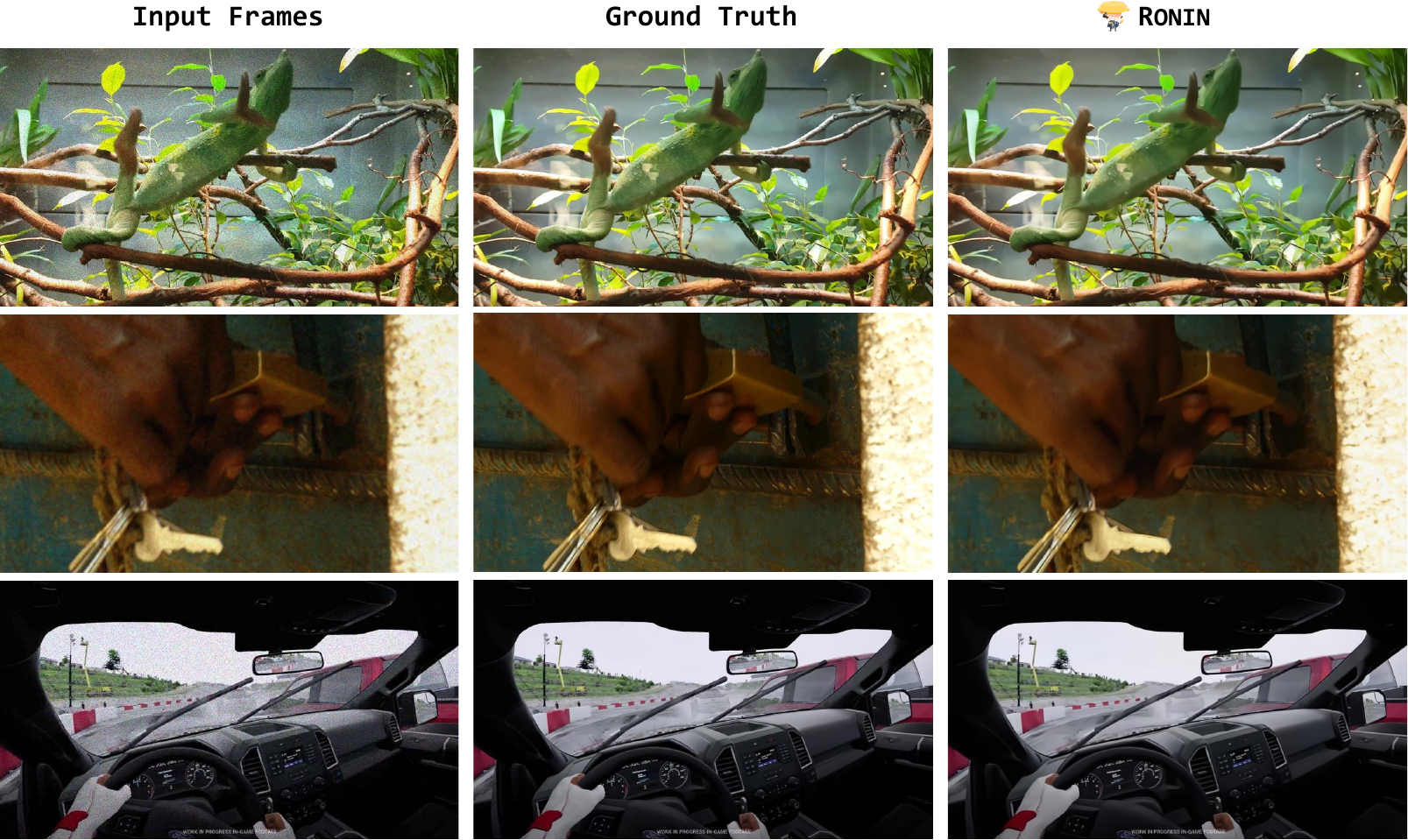}
    \caption{\textbf{TUD Benchmark Visual Results.} Qualitative results of \textsc{Ronin} on the TUD benchmark on three different settings. The first row contains frames from \(t = 6\) test set, while second and third row contains frames from \(t = 12\) and \(t = 24\) test sets, respectively. \textsc{Ronin}'s outputs are natural and faithful to the ground truth.}
    \label{fig:tud_visuals}
\end{figure*}

\begin{table}[!t]
\centering
\scalebox{1.0}{
\begin{tabular}{@{}lcccc@{}}
\toprule
\multicolumn{1}{c}{\multirow{2}{*}{\begin{tabular}[c]{@{}c@{}}\textbf{\texttt{SnowyScenes}}\\ \textbf{Statistics}\end{tabular}}} & \multicolumn{2}{c}{\textbf{GoPro}~\citep{Nah_2017_CVPR}} & \multicolumn{2}{c}{\textbf{REDS}~\citep{Nah_2019_CVPR_Workshops_REDS}} \\ \cmidrule(l){2-5} 
\multicolumn{1}{c}{} & \textbf{Train} & \textbf{Test} & \textbf{Train} & \textbf{Test} \\ \midrule
\textbf{Total Videos} & \(22\) & \(11\) & \(20\) & \(3\) \\
\textbf{Total Frames} & \(2103\) & \(1111\) & \(2000\) & \(300\) \\
\textbf{Resolution} & \multicolumn{4}{c}{\(1280\times 720\)} \\ \bottomrule
\end{tabular}
}
\caption{\textbf{Statistics of \texttt{SnowyScenes} Benchmark.} We present a summary of total videos, frames and resolution in the proposed \texttt{SnowyScenes} benchmark.}
\label{tab:snowyscenes_stats}
\end{table}

\begin{algorithm}[!t]
\caption{Prompt Algorithm}\label{alg:promptdesign}
\begin{algorithmic}
\Require Image $\mathbf{I}$
\Require Vision-Language Model $\mathbf{Q}_{\theta}$ \Comment{\textcolor{comment}{e.g., Q-Instruct}}
\State $b_{p} \gets \text{Rate the quality of the image. Think step by step.}$
\State $d_{1} \gets \mathbf{Q}_{\theta}(\mathbf{I}, b_{p})$ \Comment{\textcolor{comment}{Initial Description}}
\State $\text{desc} \gets \emptyset$
\For{$d \in \{\text{noise}, \text{rain}, ...\}$} \Comment{\textcolor{comment}{Candidate Degradations}}
\State $f_{\mathbf{I}} \gets \text{Is there}~d~\text{degradation present in the image?}$ \newline \hspace*{3.5em} $\text{Answer Yes or No.}$ \Comment{\textcolor{comment}{Fine-grained Query}}
\If{$f_{\mathbf{I}}$ is Yes}
    \State $t_{s} \gets \text{Rate the intensity of degradation}~d\text{?}$ \newline \hspace*{5.5em} $\text{Choose either severe or moderate.}$
    \State $s_{\mathbf{I}} \gets \mathbf{Q}_{\theta}(\mathbf{I}, t_{s})$ \Comment{\textcolor{comment}{Evaluate}}
    \State $d_{2} \gets \text{There is}~d~\text{in the image,}$
    \newline \hspace*{5.5em} \text{ and the intensity of}~$d$~\text{is} $s_{\mathbf{I}}$
    \State $\text{desc} \gets \textsf{concat}(d_{1}, d_{2})$ \Comment{\textcolor{comment}{Grounded Degradation}}
\EndIf
\EndFor
\end{algorithmic}
\end{algorithm}

\begin{table}[!t]
\centering
\scalebox{1.0}{
\begin{tabular}{@{}lccccc@{}}
\toprule
\textbf{Deg.} & \textbf{`Snow'} & \textbf{`Noise'} & \textbf{`Rain'} & \textbf{`Haze'} & \textbf{`Blur'} \\ \midrule
\textbf{Deblur} & \(0\) & \(1328\) & \(0\) & \(0\) & \(\mathcolor{purple}{2103}\) \\
\textbf{Derain} & \(0\) & \(5518\) & \(\mathcolor{purple}{7200}\) & \(0\) & \(1669\) \\
\textbf{Denoise} & \(6\) & \(\mathcolor{purple}{6208}\) & \(80\) & \(0\) & \(6117\) \\
\textbf{Desnow} & \(\mathcolor{purple}{26516}\) & \(13471\) & \(2\) & \(2163\) & \(10549\) \\ \bottomrule
\vspace{-2em}
\end{tabular}
}
\caption{\textbf{Robustness Analysis.} Count of degradations in the grounded degradation text from Q-Instruct~\citep{wu2023q} for the \(4\)D benchmark. The \textcolor{purple}{numbers} represent correctly classified degradations, while others are misclassifications.}
\label{tab:comparisons}
\end{table}

\section{Details of Prompting}
\label{appdx:prompttemplate}

Recall that the basic prompt to query Q-Instruct~\citep{wu2023q} to assess the degradation in the image is \textit{‘Rate the quality of the image. Think step by step.’}. While this works in most cases where the degradation matches the synthetic degradations Q-Instruct has been fine-tuned on e.g., noise, blur, brightness, clarity, it struggles to understand degradations like snow, rain, compression, and the intensity of these degradations. Therefore, we explicitly query the VLM and inquire regarding each of the candidate degradations, i.e., noise, blur, rain, compression, snow, and their appropriate combinations in the case of TUD and \texttt{SnowyScenes} benchmarks, with the answer being in a Yes/No format, while it is a multiple choice answer in the case of intensity of degradations question. A bare-bones sketch of the prompt algorithm is presented in~\cref{alg:promptdesign}. Consider a few prompt samples in~\cref{fig:more_prompt_samples}, where the first two images in the first column have moderate and severe snow, respectively, while the third image has severe noise. Also, the first image in second column has severe rain.

\subsection{Robustness of \textsc{Ronin}} We evaluate the robustness of our proposed method, \textsc{Ronin}, to misclassifications of Q-Instruct~\citep{wu2023q}. In~\cref{tab:comparisons}, we show the count of degradations \textcolor{purple}{accurately identified} by the MLLM and misclassifications. Since the dataset is video-based, naturally blur and noise (e.g., motion blur) occur, and as we lack appropriate ground truth (e.g., no blur but only snow in desnow data), we do not clean the prompts. We find that \textsc{Ronin} is robust and handles these cases well due to degradation information from the first encoder (see~\cref{fig:mainfig}), and learnable prompts initialized from latent features. Notably, degradations like snow, rain, and haze, which are not caused by camera equipment, have minimal misclassifications. For example, only \(80\) out of \(6208\) frames in the noise dataset were misidentified as rain. In the desnow data, haze was occasionally flagged, but the authors of desnow dataset~\citep{chensnow} consider snow+haze as one degradation, so we do not consider haze separately.

\section{Limitations, Future Work, and Impact}
The descriptions may occasionally include more degradations than are present in the video, such as the mention of noise in a frame which is a part of a video in the deblurring task. Although this rarely happens, as Q-Instruct~\citep{wu2023q} when prompted appropriately is adept at grounding degradations, we hypothesize that as such models improve, \textsc{Ronin} will directly benefit from their advancements. We do not correct such descriptions due to the assumption of no access to individual degradations, but improving the prompt template should also benefit \textsc{Ronin} which we leave for future work, see~\cref{fig:limitations} for few examples of such cases.

\subsection{Ethics and Societal Impact}
\label{appdx:ethics}
This work introduces a method, \textsc{Ronin}, and a benchmark dataset, \texttt{SnowyScenes}, to help advance the study of machine learning, particularly for video restoration. While the proposed method effectively restores the degraded videos, we recommend expert supervision in sensitive applications. Further, our proposed benchmark is constructed from two publicly available datasets, namely GoPro~\citep{Nah_2017_CVPR} and REDS~\citep{Nah_2019_CVPR_Workshops_REDS}. The snow is synthesized using assets of two different types of snows (for moderate and severe snow). All of the assets and both the datasets are distributed under the Creative Commons Attribution 4.0 International (CC BY 4.0) license\footnote{\href{https://creativecommons.org/licenses/by/4.0/}{https://creativecommons.org/licenses/by/4.0/}}. Therefore, \texttt{SnowyScenes} will also be distributed under the same CC BY 4.0 license.
\end{document}